\documentclass{ieeetj}
\usepackage{cite}
\usepackage{dirtree}
\usepackage{forest}
\usepackage{amsmath,amssymb,amsfonts}
\usepackage{graphicx}
\usepackage{float}
\usepackage{booktabs,tabularx,array,xcolor}
\usepackage{textcomp}
\usepackage{hyperref}
\usepackage{algorithm,algorithmic}
\usepackage{rotating}
\usepackage{threeparttable}
\usepackage{subcaption}
\newcommand{\CM}{\checkmark}
\def\BibTeX{{\rm B\kern-.05em{\sc i\kern-.025em b}\kern-.08em
    T\kern-.1667em\lower.7ex\hbox{E}\kern-.125emX}}
\AtBeginDocument{\definecolor{tmlcncolor}{cmyk}{0.93,0.59,0.15,0.02}\definecolor{NavyBlue}{RGB}{0,86,125}}
\newcommand{\X}{\textbf{x}}    
\newcolumntype{Y}{>{\raggedright\arraybackslash}X}
\renewcommand{\arraystretch}{1.15}
\renewcommand{\authorrefmark}[1]{\textsuperscript{\normalfont#1}}

\newcolumntype{C}[1]{>{\centering\arraybackslash}m{#1}} 

\def\OJlogo{}
\def\seclogo{}

\makeatletter
\def\ps@headings{%
  \def\@oddhead{}%
  \def\@evenhead{}%
  \def\@oddfoot{\hbox to \textwidth{\hfill{\rffont\thepage}\hfill}}%
  \let\@evenfoot\@oddfoot
}%
\def\ps@plain{%
  \def\@oddhead{}%
  \def\@evenhead{}%
  \def\@oddfoot{\hbox to \textwidth{\hfill{\rffont\thepage}\hfill}}%
  \let\@evenfoot\@oddfoot
}%
\AtBeginDocument{%
  \ps@headings
}
\makeatother

\begin{document}


\markboth{}{Author {et al.}}

\title{HortiMulti: A Multi-Sensor Dataset for Localisation and Mapping in Horticultural Polytunnels}

\author{Shuoyuan Xu\authorrefmark{1}, Zhipeng Zhong\authorrefmark{1}, Tiago Barros\authorrefmark{2}, Matthew Coombes\authorrefmark{1}, \\ Cristiano Premebida\authorrefmark{2}, Hao Wu\authorrefmark{3}, and Cunjia Liu\authorrefmark{1} (Member, IEEE)}
\affil{Department of Aeronautical and Automotive Engineering, Loughborough University, Leicestershire, UK, LE11 3TU}
\affil{Department of Electrical and Computer Engineering, Institute of Systems and Robotics, University of Coimbra, Coimbra, Portugal}
\affil{Antobot Ltd, Arise Innovation Hub, Chelmsford, Essex, UK, CM1 1SQ}

\begin{abstract}
Agricultural robotics is gaining increasing relevance in both research and real-world deployment. As these systems are expected to operate autonomously in more complex tasks, the availability of representative real-world datasets becomes essential. While domains such as urban and forestry robotics benefit from large and established benchmarks, horticultural environments remain comparatively under-explored despite the economic significance of this sector. To address this gap, we present HortiMulti, a multimodal, cross-season dataset collected in commercial strawberry and raspberry polytunnels across an entire growing season, capturing substantial appearance variation, dynamic foliage, specular reflections from plastic covers, severe perceptual aliasing, and GNSS-unreliable conditions — all of which directly degrade existing localisation and perception algorithms. The sensor suite includes two 3D LiDARs, four RGB cameras, an IMU, GNSS, and wheel odometry. Ground truth trajectories are derived from a combination of Total Station surveying, AprilTag fiducial markers, and LiDAR-inertial odometry, spanning dense, sparse, and marker-free coverage to support evaluation under both controlled and realistic conditions. We release time-synchronised raw measurements, calibration files, reference trajectories, and baseline benchmarks for visual, LiDAR, and multi-sensor SLAM, with results confirming that current state-of-the-art methods remain inadequate for reliable polytunnel deployment, establishing HortiMulti as a one-stop resource for developing and testing robotic perception systems in horticulture environments.
\end{abstract}



\maketitle

\section{Introduction}
Polytunnels form the backbone of soft fruit production worldwide, with protected horticulture expanding globally. China accounts for more than 80\% of the global protected area at 2.67 million hectares in 2021, Australia maintaining 13,932~ha, and India cultivating between 30,000 and 40,000~ha under protection \cite{menzel2025review}. Within soft fruit, polytunnel adoption is particularly intensive in Europe, with 97\% of the strawberry cropped area in Spain \cite{romero2020environmental} and more than 80\% in the United Kingdom \cite{Defra2024}. Yet despite this widespread adoption, the sector faces a deepening structural workforce crisis: soft fruit production is among the most labour-intensive branches of agriculture, and the supply of seasonal workers is declining across developed economies due to urbanisation, ageing rural populations, and tightening immigration policies \cite{korir2024investing, BBC2017, BSF2019}. Autonomous robotic platforms are increasingly recognised as the long-term solution to sustaining productivity in the sector \cite{bac2014harvesting, fountas2020agricultural}.

\begin{figure*}[htbp]
  \centering
  \includegraphics[width=\textwidth, trim={0cm 0cm 0cm 0cm},clip]{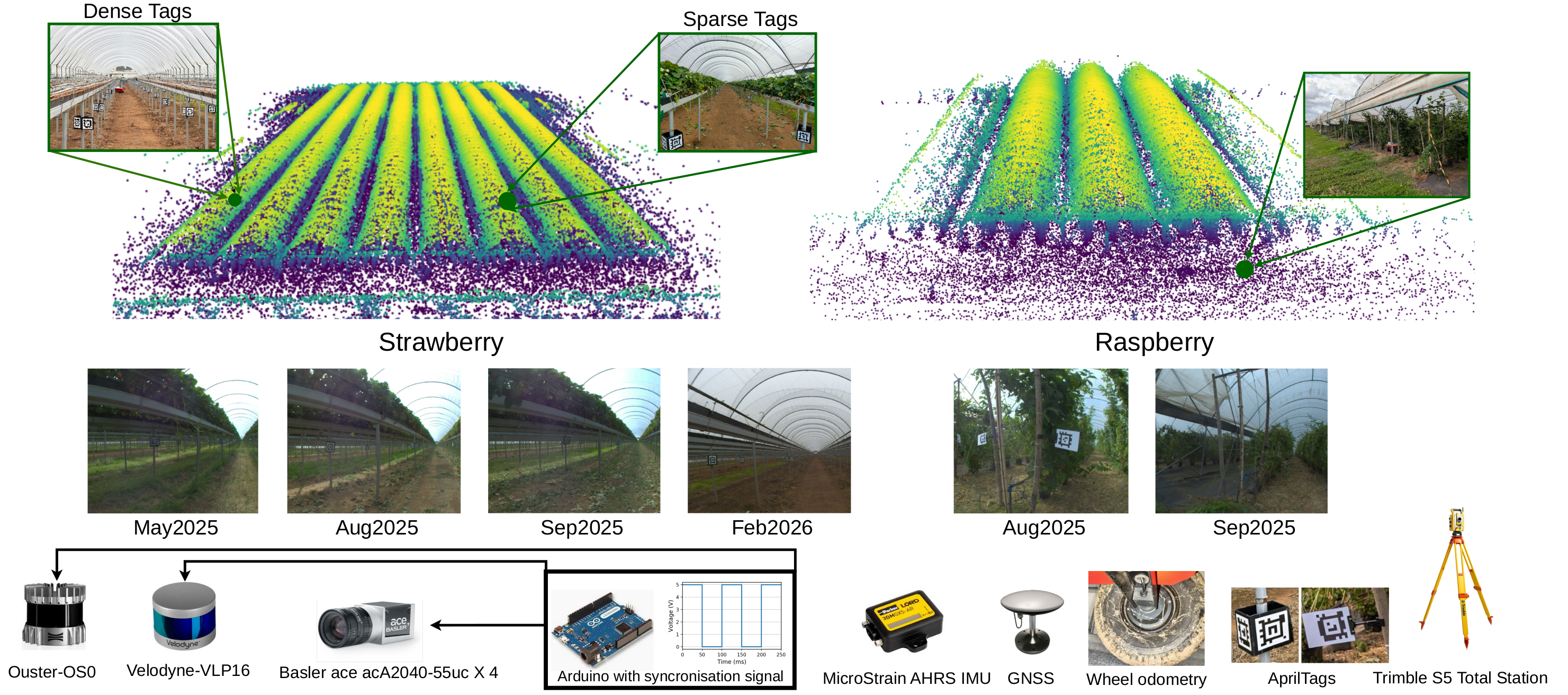}
  \caption{HortiMulti is a multi-seasonal dataset collected in strawberry (left) and raspberry (right) polytunnels. Data are synchronised across multiple LiDARs and multiple cameras; additional sensors include an AHRS, GNSS, and wheel odometry (bottom). Ground truth is obtained using AprilTags and surveying equipment (bottom right).}
  \label{fig:overview}
\end{figure*}

Deploying autonomous robots in commercial polytunnels presents navigation challenges that are fundamentally different from those encountered in the urban, indoor, and open-field environments that dominate current robotics research. Key challenges include:
\begin{itemize}
    \item \textbf{Perceptual aliasing and loop-closure ambiguity} from long, almost identical rows, rib-like tunnel frames, and regularly spaced supports or pots, which confuse place recognition \cite{Le2019,LFSD2023}.
    \item \textbf{Appearance dynamics:} wind-induced plant motion violates the static-world assumption for both LiDAR SLAM and visual SLAM, where feature repeatability across scans/frames is impaired. 
    \item \textbf{Illumination instability:}  Translucent plastic covers produce highly variable lighting, with abrupt transitions at tunnel entrances causing image saturation, undermining the photometric consistency assumptions of visual and visual-inertial odometry methods.
    \item \textbf{GNSS unreliability} from metallic structures and cover materials, requiring robust on-board localisation \cite{Das2024}.
\end{itemize}

Addressing these challenges requires representative benchmarking data. However, while the robotics community has made significant progress in developing navigation datasets for urban \cite{Geiger2013}, indoor \cite{Burri2016}, and more recently agricultural environments such as vineyards \cite{polvara2024bacchus} and orchards \cite{crocetti2023ard}, commercial polytunnels remain unrepresented in the dataset landscape. Existing agricultural datasets capture challenges such as rough terrain, open-sky illumination variability, and perceptual aliasing, yet none reproduce the combined and distinctive conditions of commercial polytunnels. Consequently, researchers developing navigation systems for polytunnel deployment lack a representative dataset to evaluate their methods, and the transferability of algorithms validated on existing datasets to polytunnel conditions remains unclear. This gap constrains progress in a domain where robust localisation and mapping are the foundational capability on which all other robotic tasks depend.

This paper addresses the lack of representative datasets for horticultural robotics by providing the first comprehensive multi-sensor, multi-season dataset specifically targeting polytunnels (Fig. \ref{fig:overview}). Our dataset captures two prevalent horticultural layouts: ground-pot systems representing traditional cultivation, exemplified here by raspberry polytunnels, and raised-table (or tabletop) systems featuring breast height beds designed for improved worker access, exemplified by strawberry polytunnels. Each presents distinct structural layouts, plant densities, and geometric characteristics, yet all share the fundamental challenge of enabling precise localisation in GNSS-unreliable, perceptually aliasing, motion varying, and illumination changing environments where conventional VSLAM and LiDAR SLAM approaches frequently fail.

HortiMulti enables systematic evaluation of SLAM and place recognition algorithms under the challenging conditions of commercial polytunnel environments. We benchmark a representative suite of state-of-the-art (SOTA) LiDAR, visual, and visual-inertial odometry methods, demonstrating that current approaches suffer from significant drift accumulation, degraded loop closure, and reduced robustness under the perceptual aliasing and illumination variability characteristic of these environments. The dataset comprises calibrated and synchronised multi-sensor sequences across multiple seasons and layouts, ground truth trajectories obtained via AprilTag-based localisation and surveying equipment, and baseline benchmark results to support reproducible evaluation by the research community.

The rest of the paper is organised in the following order: Section II reviews related datasets. Section III describes our sensor platform, synchronisation approaches, and calibration procedures. Section IV details our data collection methodology including test site and sequence collection procedure. Section V presents the ground truth generation pipeline and its corresponding accuracy analysis. Section VI presents benchmark results from SOTA SLAM algorithms and place recognition algorithms and analyses their performance characteristics in polytunnel environments. Section VII discusses key findings, future research directions and concludes the paper.

\begin{table*}[htbp]
\scriptsize
\centering
\caption{Summry of agricultural / off-road SLAM datasets.}
\label{datasetsummary}

\begin{tabularx}{\textwidth}{p{2.8cm} p{1.8cm} c c c c c c c p{4.5cm}}
\toprule
\textbf{Dataset} & \textbf{Platform} & \textbf{Lighting} & \textbf{Season} & 
\multicolumn{5}{c}{\textbf{Sensor}} & \textbf{Ground Truth} \\
\cmidrule(lr){5-9}
& & & & \textbf{LiDAR} & \textbf{Mono} & \textbf{Stereo} & \textbf{IMU} & \textbf{Wheel Odom} & \\
\midrule
\multicolumn{10}{l}{\textbf{Forest and Off-Road}} \\
\midrule

TB-Places \cite{leyva2019tb} & UGV & \X & \X & \X & \CM & \X & \CM & \CM & Laser tracker \\
Wild-Places \cite{knights2022wild} & Handheld & \X & \CM & \CM & \CM & \X & \CM & \X & SLAM\\
FinnForest \cite{ali2020finnforest} & Car & \CM & \CM & \X & \CM & \CM & \CM & \X & GNSS/INS \\
SFU Mountain \cite{bruce2015sfu} & UGV & \CM & \X & \CM & \CM & \CM & \CM & \CM & Manually-aligned locations \\
RELLIS-3D \cite{jiang2021rellis} & UGV & \X & \X & \CM & \CM & \CM & \CM & \CM & Perception labels (semantics) \\
Montmorency \cite{Tremblay2020} & Handheld/UGV & \X & \X & \X & \CM & \X & \CM & \X & Sparse GT \\
TreeScope \cite{Cheng2023TreeScope} & UAV/Backpack & \X & \X & \CM & \CM & \CM & \CM & \X & Forest inventory \\
TAIL \cite{yao2024tail} & UGV & \X & \X & \CM & \CM & \CM & \CM & \CM & RTK-GNSS \\
DiTer \cite{jeong2024diter} & Legged robot& \X & \X & \CM & \CM & \X & \CM & \X & SLAM\\
DiTer++ \cite{kim2025diter++} & Legged robot & \X & \CM & \CM & \CM & \X & \CM & \X & Terrestrial laser scanner\\
BotanicGarden \cite{liu2024botanicgarden} & UGV & \CM & \X & \CM & \CM & \CM & \CM & \X & Terrestrial laser scanner\\
ROVER \cite{schmidt2025rover} & UGV & \CM & \CM & \X & \CM & \CM & \CM & \CM & Total-station and tracker \\

\midrule
\multicolumn{10}{l}{\textbf{Agricultural}} \\
\midrule

FieldSAFE \cite{kragh2017fieldsafe} & Tractor & \X & \X & \CM & \CM & \CM & \CM & \X & RTK-GNSS \\
Sugar Beet \cite{chebrolu2017agricultural} & UGV & \X & \CM & \CM & \CM & \CM & \X & \CM & RTK-GNSS \\
Rosario \cite{pire2019rosario} & UGV & \X & \X & \X & \CM & \CM & \CM & \CM & RTK-GNSS \\
CitrusFarm \cite{teng2023multimodal} & UGV & \CM & \X & \CM & \CM & \CM & \CM & \CM & RTK-GNSS \\
Magro \cite{marzoa2024magro} & UGV & \CM & \CM & \CM & \CM & \CM & \CM & \CM & RTK-GNSS \\
Agri-Robotic \cite{polvara2022collection} & UGV & \X & \CM & \CM & \CM & \CM & \CM & \X & RTK-GNSS \\
BLT \cite{polvara2024bacchus} & UGV & \CM & \CM & \CM & \CM & \CM & \CM & \CM & RTK-GNSS \\
ARD-VO \cite{crocetti2023ard} & UGV & \CM & \CM & \CM & \CM & \CM & \CM & \X & RTK-GNSS \\
Under-Canopy \cite{cuaran2024under} & UGV & \CM & \CM & \X & \CM & \CM & \CM & \CM & RTK-GNSS (20\% coverage) \\
Greenbot \cite{canadas2024multimodal} & GV & \CM & \X & \CM & \CM & \CM & \CM & \X & No GT \\

\midrule
\multicolumn{10}{l}{\textbf{Ours}} \\
\midrule
HortiMulti & UGV & \CM & \CM & \CM & \CM & \CM & \CM & \CM & Surveyed GT and AprilTags \\

\bottomrule
\end{tabularx}
\end{table*}

\section{Related Work}

Agricultural robotics operates across diverse crop settings, each posing distinct challenges for robot localisation and mapping. Polytunnels represent a distinct environment within this space, characterised by repetitive crop rows, seasonal changes, dynamic plant motions, and GNSS-unreliable conditions — a combination that existing outdoor field and agricultural datasets only partially addressed. In the following sections, we review the most relevant existing datasets across these domains. Table \ref{datasetsummary} summarises their platform types, environment settings, sensor configurations, and availability of seasonal and ground truth data, highlighting the gaps that our dataset addresses for polytunnel robotics research.

\subsection{Forest, Off-Road, and Garden Datasets}
Several forest and off-road datasets are limited by insufficient ground truth quality. SFU Mountain \cite{bruce2015sfu}, one of the earliest multi-sensor UGV datasets in woodland trails, covers seasonal revisits but offers only limited pose ground truth. Wild-Places \cite{knights2022wild} uses handheld LiDAR–camera–IMU rigs in narrow forested trails but lacks absolute ground truth. DiTer \cite{jeong2024diter} uses LiDAR SLAM as its own reference, which limits the reliability of its ground truth. DiTer++ \cite{kim2025diter++} improves on this with a Leica RTC360 scanner and extends to multi-robot and day–night settings, though it remains focused on general terrain traversal rather than agricultural environments.

Others are limited by scenario relevance. Montmorency \cite{Tremblay2020} and TreeScope \cite{Cheng2023TreeScope} feature dense vegetation but are oriented toward forest inventory, providing no reference data suitable for navigation system development. RELLIS-3D \cite{jiang2021rellis} and TB-Places \cite{leyva2019tb} target semantics and garden settings respectively, neither of which reflects the structural or operational characteristics of polytunnels. TAIL \cite{yao2024tail} captures deformable and sandy terrain that, while challenging, shares little with row-crop environments. FinnForest \cite{ali2020finnforest} captures useful seasonal and appearance variation but in environments and with sensor configurations not representative of agricultural operation.

The most technically comparable works are ROVER \cite{schmidt2025rover} and BotanicGarden \cite{liu2024botanicgarden}. ROVER spans five semi-structured outdoor locations across all four seasons and day/dusk/night conditions, but does not include LiDAR — a primary sensor for robust outdoor navigation. BotanicGarden features UGV operation across varied natural settings with ground truth derived from a TLS-built map and scan-to-map matching. It sets a high technical standard in hardware synchronisation, sensor diversity, and ground truth quality that our dataset likewise aims to match, while targeting the specific demands of polytunnel environments.

These forest, off-road, and garden datasets address vegetation density, lighting extremes, seasonal appearance changes, and geometric degeneracy under tree canopies that partially overlap with polytunnel challenges. However, they lack the agricultural context, systematic multi-session seasonal overlaps, or extreme row-structure repetition and aliasing that define polytunnel robotics, motivating our work.

\subsection{Agricultural Datasets}
Early agricultural datasets such as FieldSAFE \cite{kragh2017fieldsafe} and the Sugar Beet dataset \cite{chebrolu2017agricultural} predate SLAM as a mainstream research focus; although both provide multi-sensor data across multiple seasons, their ground truth annotations and evaluation protocols were not designed for rigorous assessment of odometry and mapping tasks. The Rosario dataset \cite{pire2019rosario} is more SLAM-oriented, collecting mobile robot data in soybean fields under challenging conditions including lighting variation, perceptual aliasing, and partial GNSS denial, but the absence of LiDAR substantially limits its applicability for outdoor navigation development.

Vineyards and orchards share some structural characteristics with polytunnels, and several datasets target these settings. CitrusFarm \cite{teng2023multimodal} contributes dense vegetation and varied lighting but provides limited evaluations. Magro \cite{marzoa2024magro} captures repeated apple rows across seasons, but its unreliable IMU limits its application range. ARD-VO \cite{crocetti2023ard} offers a more complete vineyard benchmark with both vision- and LiDAR-based SLAM support, but absolute ground truth is unreliable due to intermittent RTK losses. Agri-Robotic \cite{polvara2022collection} and BLT \cite{polvara2024bacchus} provide long-term vineyard data with RTK ground truth spanning up to seven months of canopy evolution, but both operate in open vineyard rows that lack the enclosed structure and perceptual aliasing characteristic of polytunnels.

The environments most structurally similar to polytunnels are those of Under-Canopy \cite{cuaran2024under} and Greenbot \cite{canadas2024multimodal}. Under-Canopy covers corn, soybean, and sorghum fields with a broad multi-session field, but lacks LiDAR and provides RTK coverage for only 20\% of sequences. Greenbot is, to the best of our knowledge, the only publicly available polytunnel or greenhouse dataset, yet its limited ground truth, restricted variety of tunnel types, and narrower sensor suite reduce its practical value for polytunnel robotics development — the gap our dataset is designed to fill.

These agricultural datasets establish important foundations for perception, localisation, mapping, and long-term operation in crop environments. They demonstrate seasonal vegetation dynamics, row-structure challenges, and GNSS-degraded conditions that are highly relevant to polytunnel robotics. However, they exhibit critical limitations: (1) most focus on open-field or vineyard/orchard scenarios rather than GNSS-unreliable environments; (2) multi-session path overlaps are limited, restricting loop closure/re-localisation testing; (3) ground truth quality is mixed,  particularly where GNSS is unreliable; (4) extreme geometric repetition and aliasing characteristic of polytunnel rows are under-represented. These gaps motivate our contribution: a polytunnel-specific dataset with survey-grade ground truth, systematic multi-session overlaps, heterogeneous sensor modalities, and explicit focus on row-structure aliasing and loop-closure challenges.

\section{Platform}
This section details the robotic platform and sensor configuration used throughout the experimental evaluation, covering time synchronisation, calibration procedures, and the generation of colourised point cloud data. The sensor configuration is illustrated in Fig.~\ref{fig:RobotConfig}, with platform dimensions and sensor specifications summarised in Tables~\ref{tab:platform_geometry} and~\ref{tab:sensor_specs} respectively.

\subsection{Sensor Setup}

\begin{figure}[htbp]
  \centering
  \includegraphics[width=\columnwidth, trim={0cm 3cm 5cm 1cm},clip]{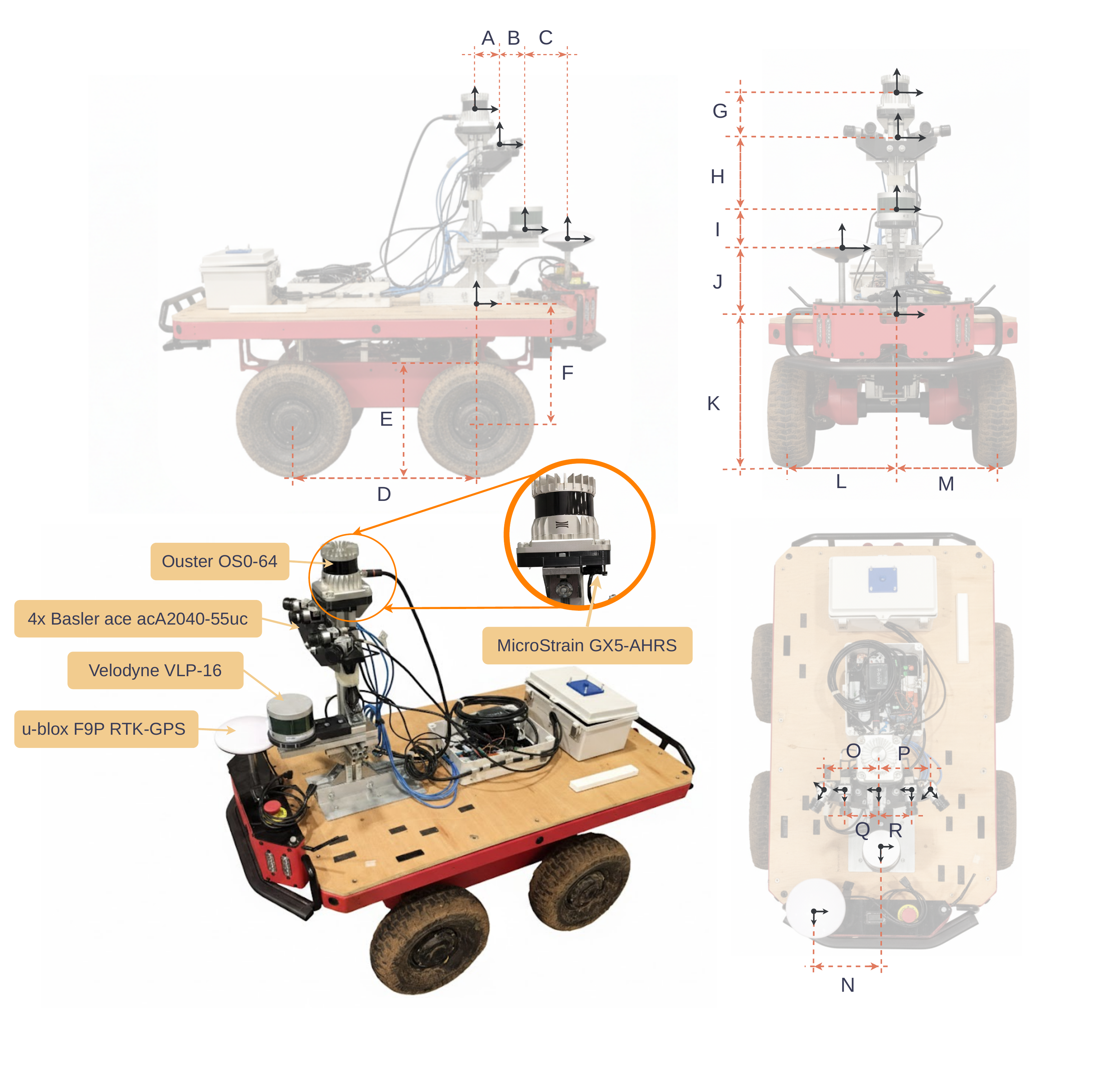}
  \caption{Robot platform and sensor tower}
  \label{fig:RobotConfig}
\end{figure}

\begin{table}[h]
\centering
\caption{Geometric parameters of the mobile sensing platform (labels refer to Fig.~\ref{fig:RobotConfig})}
\label{tab:platform_geometry}
\begin{tabular}{C{0.06\textwidth} C{0.06\textwidth} | C{0.06\textwidth} C{0.06\textwidth} | C{0.06\textwidth} C{0.06\textwidth} }
\hline
\textbf{Symbol} &  \textbf{Dim. (mm)} &\textbf{Symbol} &  \textbf{Dim. (mm)} & \textbf{Symbol} &  \textbf{Dim. (mm)} \\
\hline
A &  48  & G & 121  & M &  303\\
B &  51  & H & 199 & N &  160 \\
C &  210 & I &  43 & O &  113 \\
D & 573  & J &  197 & P &  108\\
E & 330  & K &  505 & Q &  63\\
F & 340  & L &  305 & R &  76\\
\hline
\end{tabular}
\end{table}

Our data were collected using a commercial four-wheel drive robot platform from Antobot Ltd, UK. The base robot is equipped with onboard wheel odometry, GNSS, an onboard router for external communication, and a Jetson AGX Xavier for basic computation and control.

\begin{table*}[h]
\centering
\caption{Sensor Specifications}
\label{tab:sensor_specs}
\begin{tabular}{@{}lll@{}}
\toprule
\textbf{Sensor} & \textbf{Model} & \textbf{Specification} \\ \midrule
Spinning LiDAR & Velodyne VLP-16 & 16 channels, 360$^\circ$ $\times$ 40$^\circ$ FOV, 200\,m range, 10\,Hz, $\pm3$\,cm accuracy \\
Spinning LiDAR & Ouster OS0-64 & 64 channels, 360$^\circ$ $\times$ 90$^\circ$ FOV, 100\,m range, 10\,Hz, $\pm0.8$\,cm accuracy \\
IMU & MicroStrain 3DM-GX5-AHRS & 3-axis gyro, accel, mag, 200\,Hz, $\pm0.05^\circ$ roll/pitch, $\pm0.1^\circ$ yaw \\
RGB Camera & Basler ace U acA2040-55uc & 3.2\,MP, 1936$\times$1216 px, 10\,fps global shutter \\
GNSS & u-blox F9P GPS & Multi-band GNSS, RTK $<$0.02\,m (RTK fix), 8\,Hz output \\
Wheel Odometry & Encoders & Differential drive, resolution $<$1\,mm per tick \\ \bottomrule
\end{tabular}
\end{table*}

To integrate additional sensors, we designed and manufactured a custom sensor tower using aluminium profiles and 3D printed parts, as shown in Fig. \ref{fig:RobotConfig}. This sensor tower is equipped with an Onyx NUC as the primary computational unit for sensor data processing and recording. The NUC runs Ubuntu 20.04 and is configured with an Intel Core i5-13500H CPU, 2×16 GB DDR5 RAM, a 2TB Samsung 990 Pro NVMe SSD, 3× USB 3.2 Gen2 Type-A ports, 3× USB 4.0 Type-C ports, and an Intel 2.5Gb LAN port. Communication between the sensor tower and base robot is established via Ethernet.

The platform integrates a complementary suite of sensors covering vision, LiDAR, and inertial modalities. The vision system comprises four Basler ace U acA2040-55uc colour cameras (3~MP, 10~fps): two of them configured as a forward-facing stereo pair with a 140~mm baseline, and two angled at $45^\circ$ to the sides to extend lateral field of view, each equipped with a 3.5~mm C-Mount wide-angle lens and a 1/1.8" sensor with 3.45~\textmu m pixel pitch. The cameras use hardware trigger for synchronisation and stream data via USB 3.0. The forward-facing stereo pair remains active throughout all sequences; the side-facing cameras, however, are not always enabled, as their lateral field of view provides little perceptual benefit when traversing narrow polytunnel rows.

To support different testing demands and performance evaluation, two LiDARs are employed: an Ouster OS0-64 and a Velodyne VLP-16. The OS0-64 is a high-performance spinning LiDAR with 64 beams, 360$^\circ$ $\times$ 90$^\circ$ Field-of-View (FoV), and 0–100 m range, suitable for evaluating the performance limits of existing algorithms. The VLP-16 is a cost-effective 16-beam spinning LiDAR with 360$^\circ$ $\times$ 40$^\circ$ FoV, similar effective range, and worse accuracy, enabling assessment of algorithm performance with more affordable sensors. Both LiDARs are configured to scan at 10 Hz. For inertial measurements, we employ a MicroStrain 3DM-GX5-AHRS IMU, which provides high-frequency data at 200 Hz. Detailed specifications of all sensors are listed in Table \ref{tab:sensor_specs}. Our dataset and system configuration are available on GitHub.\footnote{\url{https://github.com/shuoyuanxu/HortiMulti}} General dimensions of our platform can be found in Table \ref{tab:platform_geometry}, detailed sensor extrinsic please refer to the calibration section.

\subsection{Time Synchronisation}
In a precision robotic system with rich sensors and multiple computational hosts, time synchronisation is critical. Our architecture addresses this at two levels: hardware trigger synchronisation across sensors, and GPS clock alignment across computers. 

The hardware triggering is implemented using an Arduino Uno microcontroller, which is programmed to produce a square wave signal at 10 Hz to synchronise the LiDARs and cameras. When the rising edge of the square wave arrives, the cameras immediately begin exposure with their global shutters. The exposure duration is configured such that image acquisition completes before the falling edge, ensuring that each camera frame corresponds to a single trigger pulse. The timestamp for each image is determined by the trigger arrival time plus half the exposure duration. Simultaneously, the 10 Hz pulse signal is sent to both LiDARs (Ouster OS0-64 and Velodyne VLP-16) to phase lock them to each other.

For absolute time reference, we employ a u-blox F9P GNSS receiver, which provides high-precision GPS time with PPS (Pulse Per Second) output. The system clocks on all computational hosts—including the base robot's Jetson AGX Xavier and the sensor tower's NUC—are synchronised using Chrony NTP daemon in combination with gpsd, which set the system clocks to GPS time from the F9P receiver. This approach ensures that all sensor measurements across different hosts share a common absolute time reference.

\subsection{Calibration}
Calibration, encompassing both intrinsic and extrinsic parameters, is a prerequisite for multi-sensor fusion and algorithm development. The overall calibration quality is shown in Table \ref{tab:SensorCalibrationPrecision}.

\begin{table}[H]
\centering
\caption{Calibration quality summary}
\label{tab:SensorCalibrationPrecision}
\begin{tabular}{lll}
\toprule
\textbf{Calibration} & \textbf{Precision} & \textbf{Tool} \\
\midrule
Camera to Camera & $\leq$ 0.33 px reprojection error & \cite{rehder2016extending}\\
Camera to IMU & 0.52 px reprojection error & \cite{rehder2016extending, AllanVarianceRos}\\
Camera to LiDAR & 0.943 px reprojection error & \cite{koide2023general} \\
LiDAR to LiDAR & fitness: 0.999 & \cite{kulmermfi2024}\\
\bottomrule
\end{tabular}
\end{table}

\textbf{Camera} intrinsics and camera-to-camera extrinsics are calibrated using the Kalibr toolbox \cite{rehder2016extending} with a pinhole model and equidistant distortion. A Kalibr target board is presented at various distances and orientations to the camera array, achieving a mean reprojection error below 0.33 pixels across all cameras.

\textbf{IMU} noise densities and random walk parameters are first characterised via Allan Variance ROS \cite{AllanVarianceRos} using 24 hours of recorded measurements. Camera-IMU extrinsics are then estimated using Kalibr with a 10$\times$10 AprilTag grid, achieving a mean reprojection error of 0.52 pixels.

\textbf{Camera-to-LiDAR} extrinsics are estimated using the direct visual-LiDAR calibration library \cite{koide2023general}, with the left camera and Ouster OS0-64 as reference frames. Calibration is performed by selecting 20+ point correspondences between the camera image and an ambient point cloud, achieving a reprojection error of 0.943 pixels. The result is validated by projecting colourised LiDAR points onto the image plane, as shown in Fig.~\ref{fig:colourPC}.

\textbf{LiDAR to LiDAR} extrinsic between the Velodyne VLP-16 and Ouster OS0-64 is obtained using Multi-LiCa \cite{kulmermfi2024}, a fully automatic calibration toolbox requiring no calibration targets or manual point selection. It combines FPFH feature extraction, TEASER++ coarse alignment, and GICP refinement on stationary point clouds collected simultaneously from both LiDARs, achieving a registration fitness of 99.94\%.

\subsection{Colourised point cloud}
\begin{figure}[htbp]
  \centering
  \includegraphics[width=0.9\linewidth]{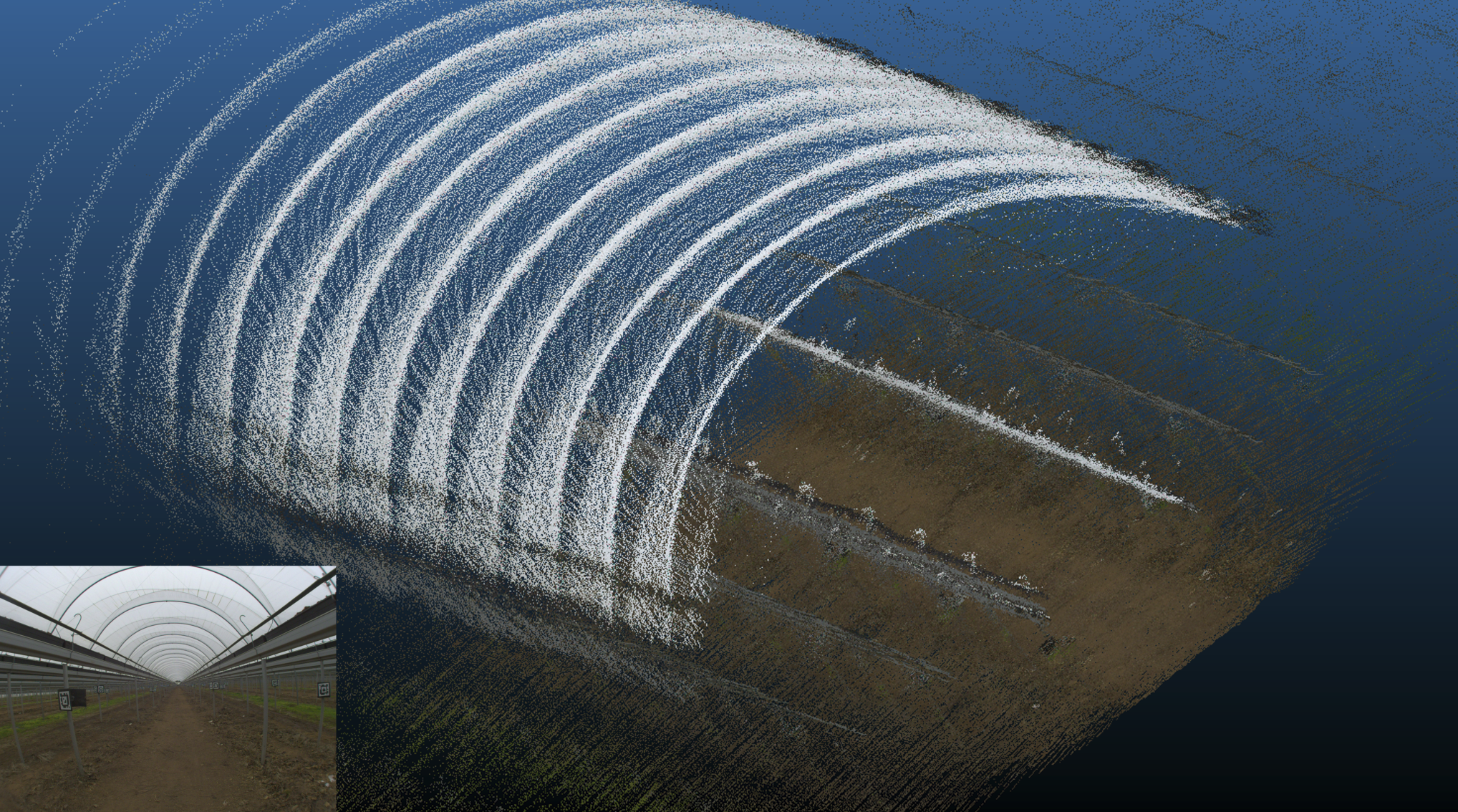}
  \caption{Example of a colourised point cloud with corresponding image (bottom left). }
  \label{fig:colourPC}
\end{figure}

A unique feature of our dataset is colourised point clouds, where RGB information from the side camera pair is projected onto the Ouster OS0-64 LiDAR point cloud. This process leverages the calibrated camera intrinsics and camera-LiDAR extrinsics to rectify and register the images to the 3D point cloud. An example of the resulting coloured point cloud is shown in Fig. \ref{fig:colourPC}. This RGB-enriched 3D representation provides complementary information for learning-based perception tasks, particularly for semantic segmentation and object recognition in agricultural environments. The colourisation tool is publicly available with the dataset.\footnote{\url{https://github.com/shuoyuanxu/Point_Cloud_RGB_colorising}}

\section{Data Collection}
This section presents the environment, the data recording protocol, and the dataset structure. We first provide an overview of the environments in which the dataset was collected, as shown in Fig.~\ref{fig:overview}. We then describe the recording methodology, illustrated in Fig.~\ref{fig:trajectoryoverview}. Finally, we detail the dataset structure, depicted in Fig.~\ref{fig:datastructure}, and summarise the recorded topics in Table~\ref{tab:topics}.

\subsection{Polytunnel Environment}
The experiments were conducted at the Haygrove facilities in Newent, Herefordshire, UK, which were selected for their wide range of commercially representative polytunnel systems. Two predominant industry-standard configurations were studied: strawberry polytunnels and raspberry polytunnels.

The strawberry polytunnel data was collected across eight polytunnels (51\textdegree56'N, 2\textdegree23'W), each measuring 160\,m in length and 10\,m in width. Each tunnel featured a 2\,m-wide central aisle running the full length, which served as the primary access route for agricultural machinery. Secondary pathways, approximately 1\,m wide, extended laterally from this central aisle to provide access for growers and fruit pickers. The cultivation system comprised six rows of strawberry plants arranged on metal tabletop structures, elevated to breast height (1.5\,m) by steel poles to facilitate ergonomic harvesting. This configuration represents a highly structured polytunnel environment with well-defined geometric features.

\begin{figure}[!tp]
  \centering
  \includegraphics[width=0.9\linewidth]{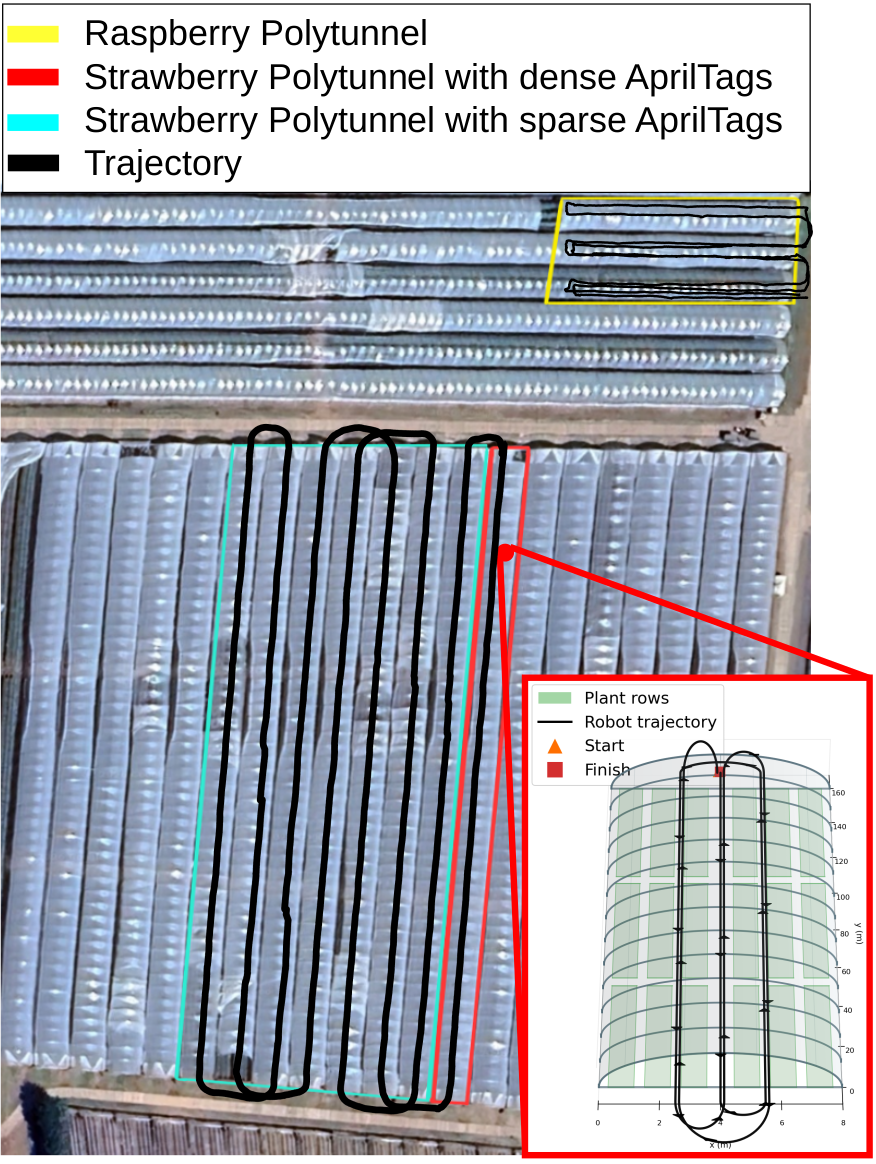}
  \caption{Example of some representative trajectories, designed to mimic real farm machinery operations: cross-tunnel passes that mimic spraying tasks, and row-wise traverses within a single polytunnel that resemble UV treatment and potential harvesting robots.}
  \label{fig:trajectoryoverview}
\end{figure}

In contrast, the raspberry experiments were conducted in six ground-based polytunnels, each 120 m long and 10 m wide, located approximately 50 m from the strawberry tunnels (see Fig. \ref{fig:trajectoryoverview} for details). Raspberry plants were grown in pots placed directly on the ground, arranged in rows with approximately 1 m spacing. During the growing season, plant canopy coverage was dense, with raspberry foliage occupying most of the observable space within the tunnel, including the sidewalls and roof area. This setup represents a less structured, vegetation-dominated polytunnel environment.

\subsection{Recording Overview}
Our dataset was collected across multiple time periods spanning June 2025 to February 2026, covering the full seasonal cycle from growing season through harvest to off-season. This temporal range captures the characteristic environmental variations of commercial polytunnels: evolving plant densities, diverse lighting conditions, and changes in ground surface appearance. In total, 20 sequences were collected across strawberry and raspberry polytunnels, comprising over 11~km of trajectories. The sequences encompass a broad range of operating conditions, including short single-tunnel traversals and extended multi-tunnel runs, sunny and overcast lighting, loop closures, sharp $180^\circ$ turns at tunnel ends, and highly repetitive geometric structures. Fig.~\ref{fig:trajectoryoverview} shows three representative sequences from the collection. HortiMulti is designed as a living dataset: we intend to continue expanding the collection with additional sequences, sensor configurations, and crop environments, and welcome community contributions. Benchmark results and evaluation tools will be updated accordingly as new data become available.

\subsection{Dataset Structure}
The dataset includes recordings from multiple polytunnels and dates (see Table~\ref{tab:topics}). All data are provided as ROS bag files, enabling direct integration with ROS-based SLAM and perception frameworks without requiring additional preprocessing or conversion steps. Each bag file contains the following ROS topics with their corresponding message types and approximate publish rates (Table \ref{tab:topics}):

\begin{table*}[t]
\centering
\caption{Topics stored in each data session.}
\label{tab:topics}
\begin{tabular}{llll}
\toprule
\textbf{Sensor} & \textbf{Topic Name} & \textbf{ROS Message Type} & \textbf{Rate} \\
\midrule
Ouster LiDAR        & \texttt{/ouster/points}                       & \texttt{sensor\_msgs/PointCloud2}             & $\sim$10 Hz  \\
Velodyne LiDAR      & \texttt{/velodyne\_points}                    & \texttt{sensor\_msgs/PointCloud2}             & $\sim$10 Hz  \\
Forward-Left Camera & \texttt{/forwardLeft/image\_raw/compressed}   & \texttt{sensor\_msgs/CompressedImage}         & $\sim$10 Hz  \\
 & \texttt{/forwardLeft/camera\_info}            & \texttt{sensor\_msgs/CameraInfo}              & $\sim$10 Hz  \\
 & \texttt{/forwardLeft/tag\_detections}         & \texttt{apriltag\_ros/AprilTagDetectionArray} & $\sim$10 Hz  \\
Forward-Right Camera & \texttt{/forwardRight/image\_raw/compressed} & \texttt{sensor\_msgs/CompressedImage}         & $\sim$10 Hz  \\
 & \texttt{/forwardRight/camera\_info}          & \texttt{sensor\_msgs/CameraInfo}              & $\sim$10 Hz  \\
 & \texttt{/forwardRight/tag\_detections}       & \texttt{apriltag\_ros/AprilTagDetectionArray} & $\sim$10 Hz  \\
Left Camera & \texttt{/Left/image\_raw/compressed}   & \texttt{sensor\_msgs/CompressedImage}         & $\sim$10 Hz  \\
 & \texttt{/Left/camera\_info}            & \texttt{sensor\_msgs/CameraInfo}              & $\sim$10 Hz  \\
 & \texttt{/Left/tag\_detections}         & \texttt{apriltag\_ros/AprilTagDetectionArray} & $\sim$10 Hz  \\
Right Camera & \texttt{/Right/image\_raw/compressed} & \texttt{sensor\_msgs/CompressedImage}         & $\sim$10 Hz  \\
 & \texttt{/Right/camera\_info}          & \texttt{sensor\_msgs/CameraInfo}              & $\sim$10 Hz  \\
 & \texttt{/Right/tag\_detections}       & \texttt{apriltag\_ros/AprilTagDetectionArray} & $\sim$10 Hz  \\
IMU                 & \texttt{/ms/imu/data}                         & \texttt{sensor\_msgs/Imu}                     & $\sim$200 Hz \\
GPS                 & \texttt{/antobot\_gps}                        & \texttt{sensor\_msgs/NavSatFix}               & $\sim$8 Hz   \\
Wheel Odometry          & \texttt{/antobot\_robot/odom}                 & \texttt{nav\_msgs/Odometry}                   & $\sim$25 Hz  \\
Controller Command Velocity          & \texttt{/antobot\_robot/cmd\_vel}             & \texttt{geometry\_msgs/Twist}                 & $\sim$25 Hz  \\
TF                  & \texttt{/tf}                                  & \texttt{tf2\_msgs/TFMessage}                  & ---          \\
                  & \texttt{/tf\_static}                          & \texttt{tf2\_msgs/TFMessage}                  & ---          \\
\bottomrule
\end{tabular}
\end{table*}

Camera intrinsic parameters are embedded within the \texttt{sensor\_msgs/CameraInfo} topics accompanying each image stream, following standard ROS conventions. Extrinsic calibrations between all sensors are provided through the \texttt{tf} tree, enabling automatic spatial transformations throughout the sensor suite. Calibration files are also provided separately on our website.

\begin{figure}[h]
\centering
\begin{forest}
  for tree={
    font=\ttfamily\small,
    grow'=0,
    child anchor=west,
    parent anchor=south,
    anchor=west,
    calign=first,
    edge path={
      \noexpand\path [draw, \forestoption{edge}]
      (!u.south west) +(7.5pt,0) |- (.child anchor)
      \forestoption{edge label};
    },
    before typesetting nodes={
      if n=1{insert before={[,phantom]}}{}
    },
    fit=band,
    before computing xy={l=15pt},
    edge={thin},
    inner sep=2pt,
  }
[/Dataset
  [Feb2026/
    [Strawberry\_1/
      [{\texttt{Strawberry\_1.bag}}]
      [{\texttt{GT\_trajectory.csv}}]
      [{\texttt{GT\_tag.csv}}]
    ]
    [{$\vdots$}, edge={draw=none}]
    [Raspberry\_1/
      [{\texttt{Raspberry\_1.bag}}]
      [{\texttt{GT\_trajectory.csv}}]
      [{\texttt{GT\_tag.csv}}]
    ]
    [{\texttt{calibration.yaml}}]
    [{$\vdots$}, edge={draw=none}]
  ]
  [Sep2025/
  ]
  [{$\vdots$}, edge={draw=none}]
]
\end{forest}
\caption{File structure of the dataset.}
\label{fig:datastructure}
\end{figure}
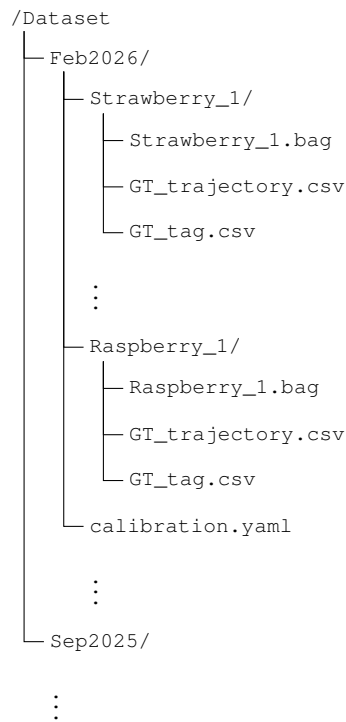

The dataset is publicly available on our project website, with the file organisation illustrated in Fig.~\ref{fig:datastructure}. Sequences are organised by recording time (e.g., \texttt{Feb2026/}) and sequence name (e.g., \texttt{Strawberry-1/}). Each sequence folder contains three files: a ROS bag file (\texttt{*.bag}); a ground truth trajectory file (\texttt{GT\_trajectory.csv}) storing timestamped poses as translation $(x, y, z)$ and quaternion $(q_w, q_x, q_y, q_z)$, aligned to ROS bag header time for direct trajectory comparison; an AprilTag position file (\texttt{GT\_tag.csv}) providing ground truth tag locations for mapping evaluation; and a calibration file (\texttt{calibration.yaml}), which is provided for each recording session rather than globally, as sensor calibration may drift over extended periods or change following minor hardware adjustments or reconfigurations of the platform between deployments. Sequences collected across different recording sessions may reflect minor variations in sensor configuration or mounting; the calibration file provided within each session folder corresponds precisely to the hardware configuration used during that recording, and users should take care to apply the correct calibration when combining sequences across sessions.

\section{Ground Truth}
Our ground truth generation strategy employs two complementary components: a Trimble S5 Total Station to establish millimetre-accurate positions for some of the AprilTag landmarks, and our Poly-TagSLAM algorithm~\cite{xu2025poly} to estimate robot trajectories by fusing tag detections, surveyed tag positions, and LiDAR-inertial odometry. Because permanent tag installation is not viable in all environments, we adopt two ground truth regimes across the dataset: survey-constrained Poly-TagSLAM, where total-station-surveyed landmarks are incorporated as tight priors, is used for strawberry polytunnel sequences where tags are permanently installed; for raspberry sequences where permanent installation is impractical, Poly-TagSLAM without survey constraints provides sufficient accuracy for the majority of evaluation tasks. The following subsections describe the tag deployment strategy, the total station surveying procedure, the pose estimation pipeline, and a rigorous quality assessment of the resulting ground truth through leave-one-out cross-validation.

\subsection{AprilTag Deployment}
We deployed AprilTags from the \texttt{TagStandard41h12} family with a physical tag size of $75 \times 75\,\mathrm{mm}$ across both strawberry and raspberry polytunnels. All tags were configured to achieve a detection range of up to $6\,\mathrm{m}$ under typical polytunnel lighting conditions.

\begin{figure}[htbp]
\centering
    \begin{subfigure}[b]{0.49\linewidth}
    \centering
    \includegraphics[width=\linewidth]{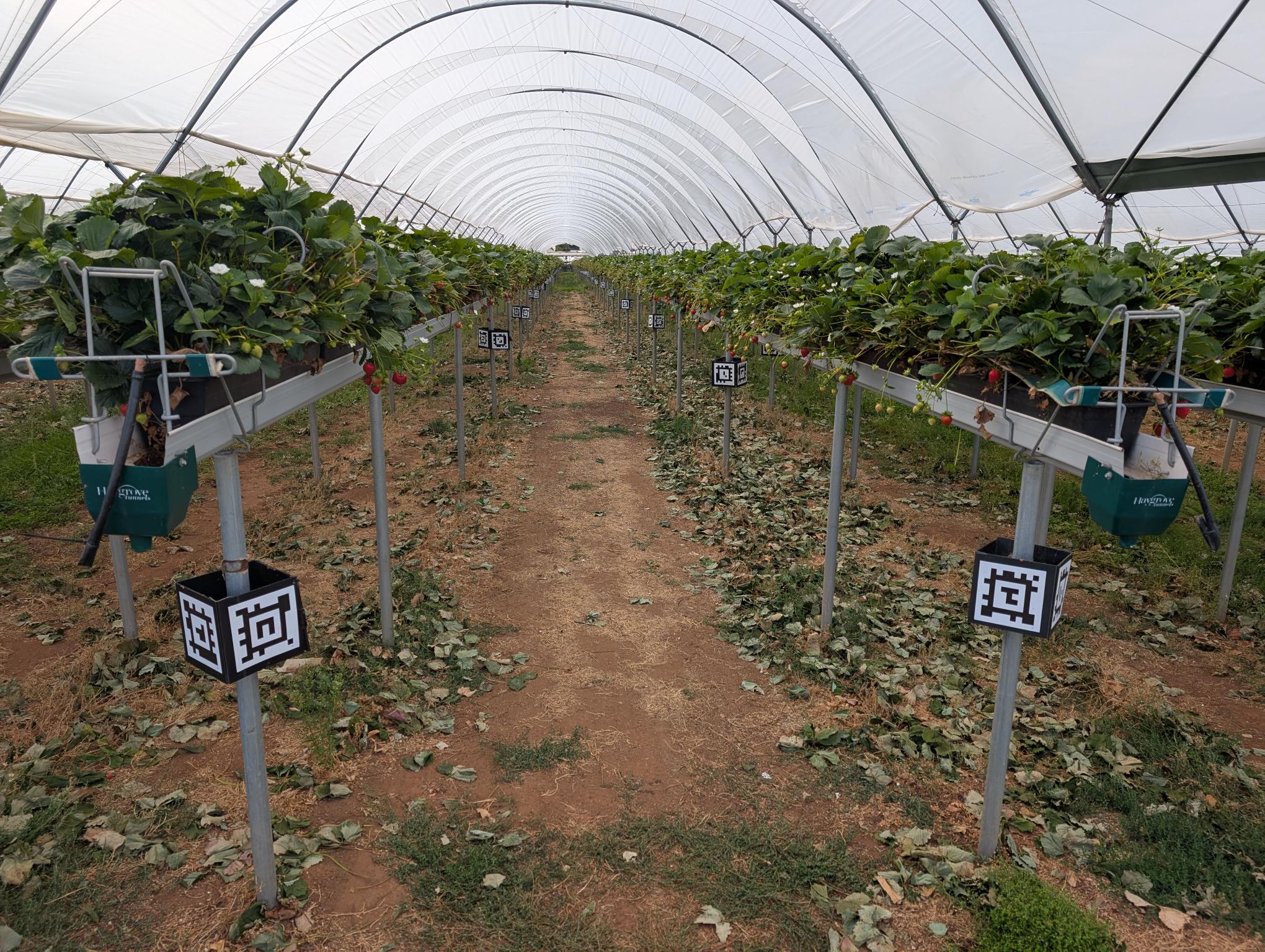}
    \caption{}
    \label{fig:strawberry_tags}  
    \end{subfigure}
    \begin{subfigure}[b]{0.49\linewidth}
    \centering
    \includegraphics[width=\linewidth]{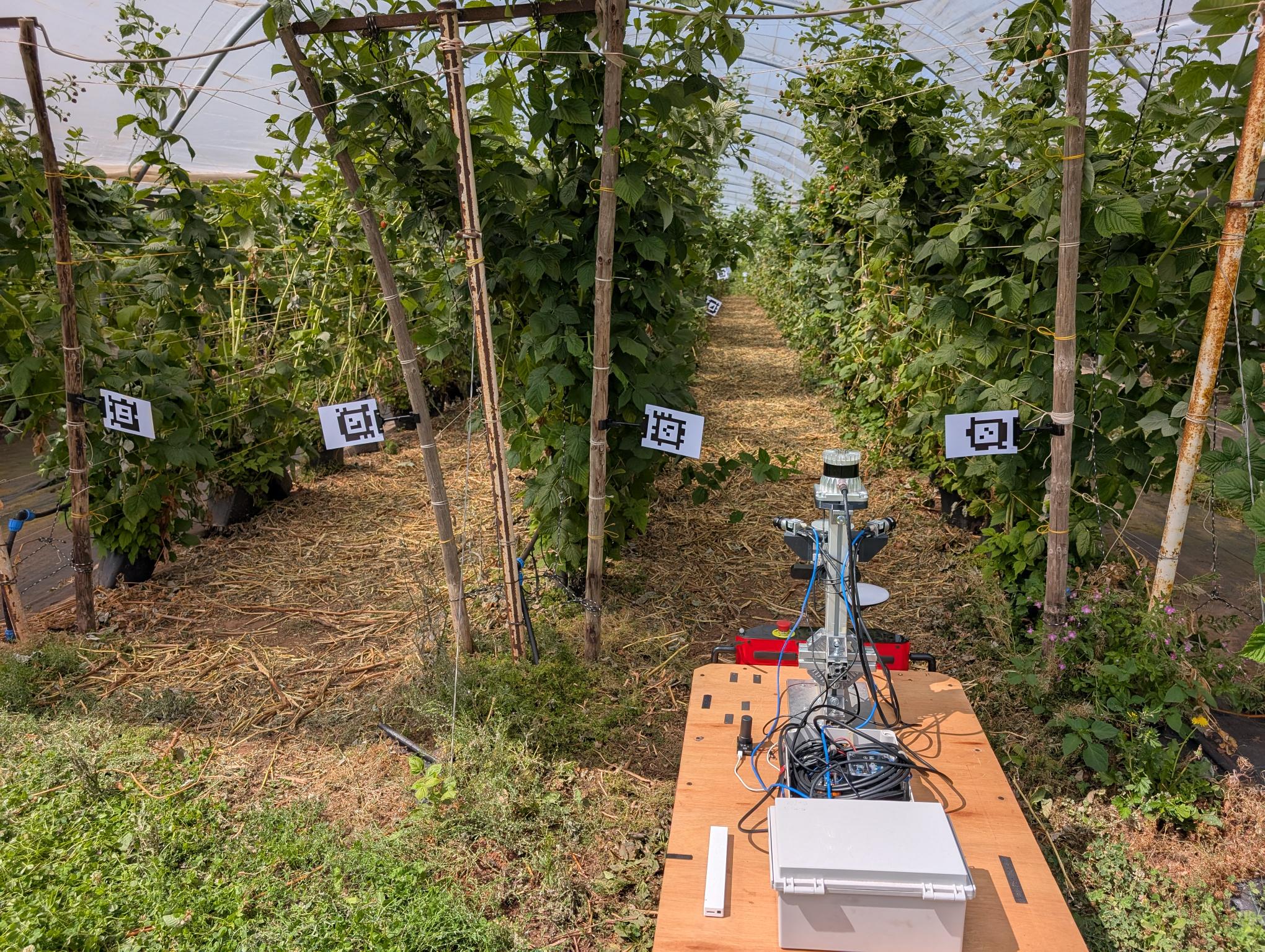}
    \caption{}
    \label{fig:raspberry_tags}
    \end{subfigure}
    
    \begin{subfigure}[b]{0.24\linewidth}
    \centering
    \includegraphics[height=3.5cm]{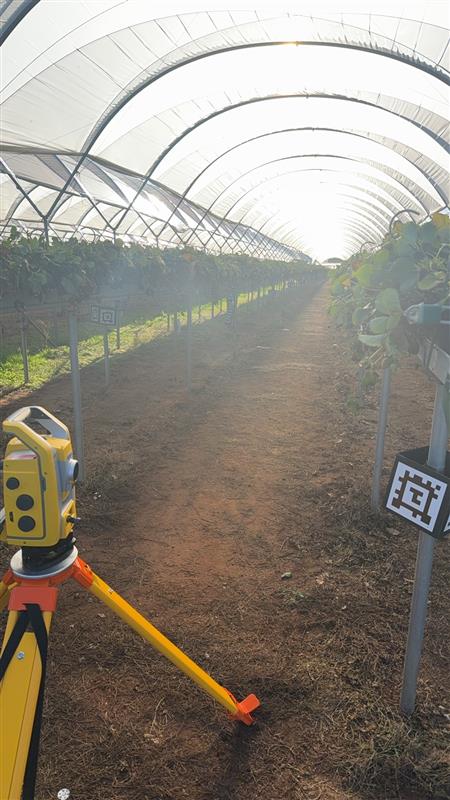}
    \caption{}
    \end{subfigure}
    \begin{subfigure}[b]{0.24\linewidth}
    \centering
    \includegraphics[height=3.5cm]{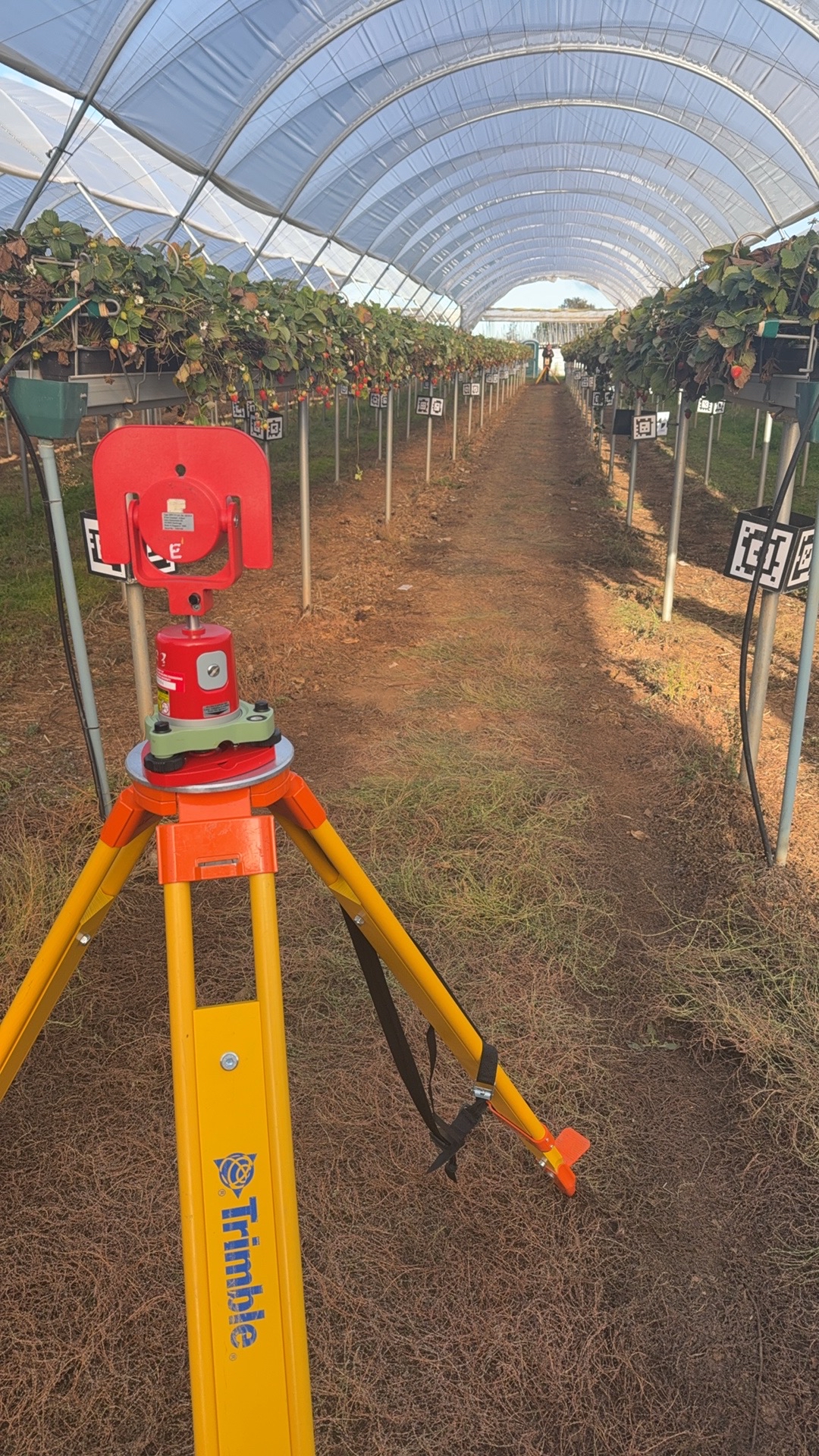}
    \caption{}
    \end{subfigure}
    \begin{subfigure}[b]{0.24\linewidth}
    \centering
    \includegraphics[height=3.5cm]{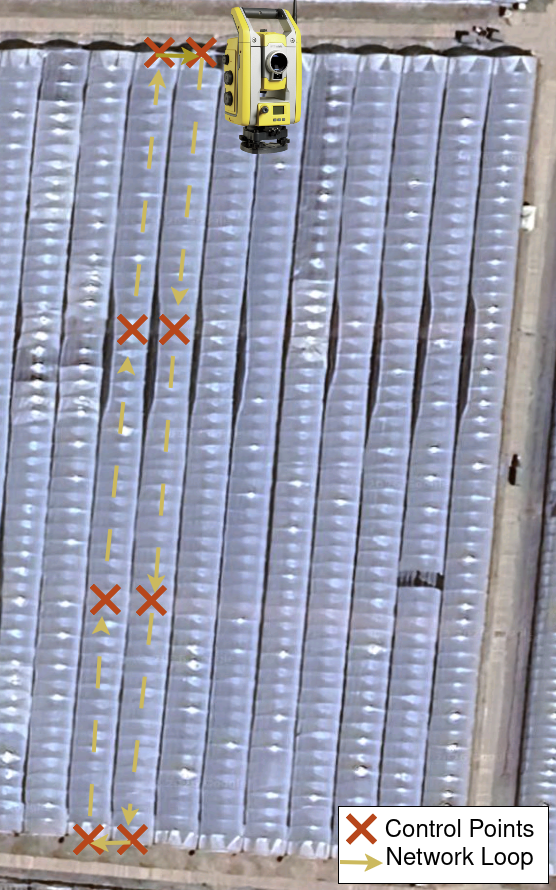}
    \caption{}
    \end{subfigure}
    \begin{subfigure}[b]{0.24\linewidth}
    \centering
    \includegraphics[height=3.5cm]{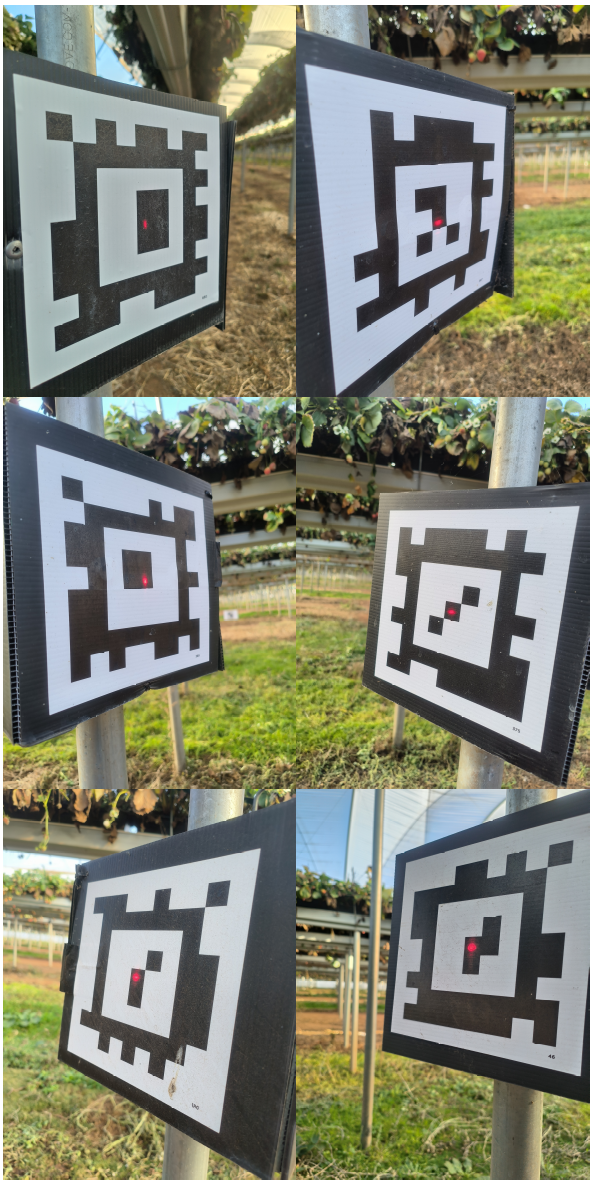}
    \caption{}
    \end{subfigure}

    \begin{subfigure}[b]{0.32\linewidth}
    \centering
    \includegraphics[height=4.7cm]{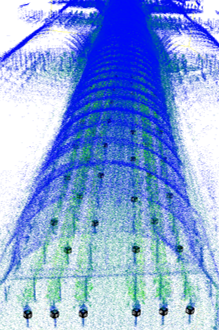}
    \caption{}
    \end{subfigure}
    \begin{subfigure}[b]{0.32\linewidth}
    \centering
    \includegraphics[height=4.7cm]{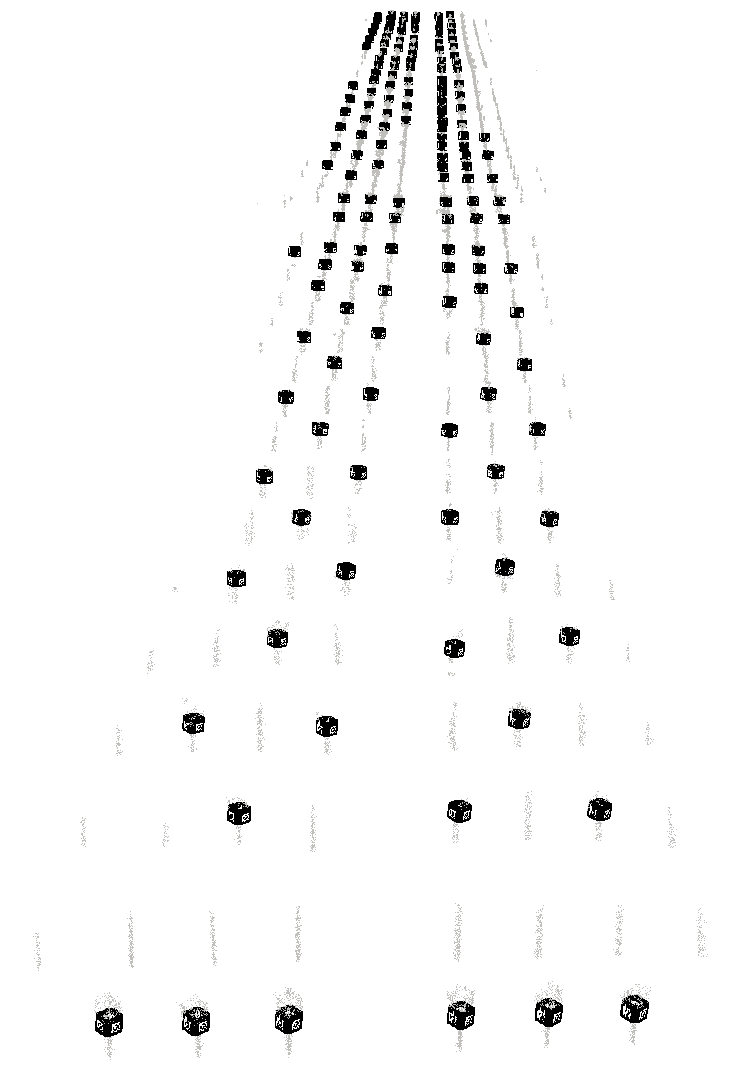}
    \caption{}
    \end{subfigure}
    \begin{subfigure}[b]{0.32\linewidth}
    \centering
    \includegraphics[height=4.7cm]{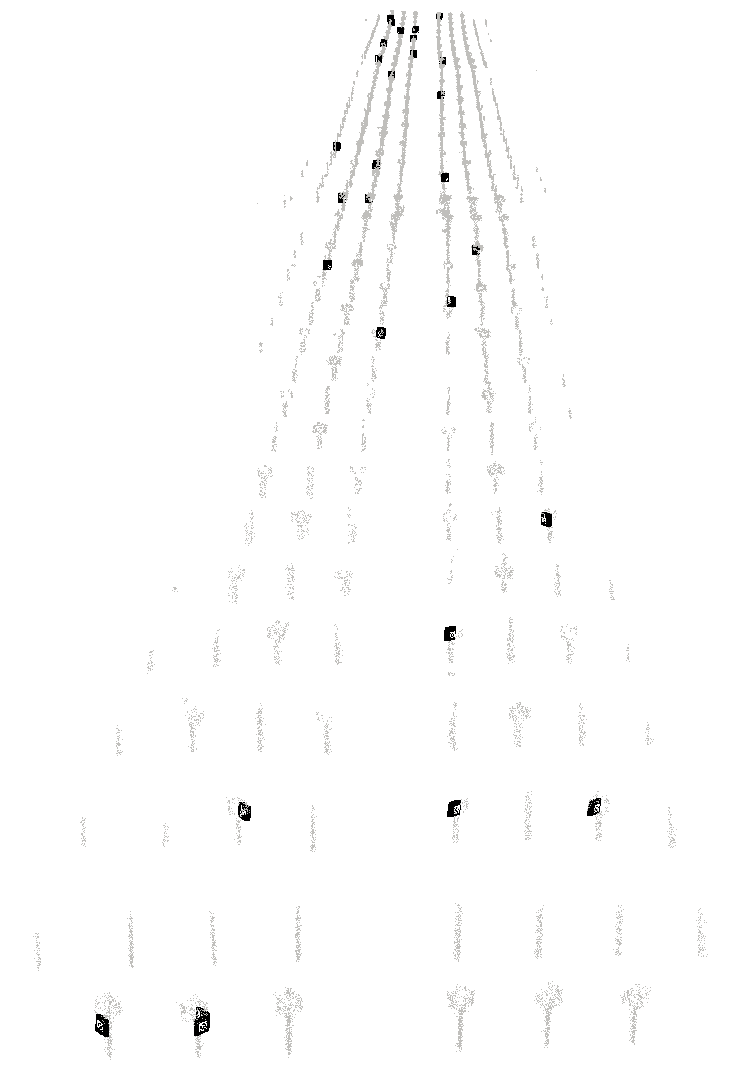}
    \caption{}
    \end{subfigure}
    
\caption{\textbf{Ground truth data collection:} (a) and (b) illustrate AprilTag installations in strawberry and raspberry polytunnels, respectively; (c)--(f) show the total station and prism setup used to establish a control network and survey the AprilTag positions; (g)--(i) present representative survey results, with (g) showing all tags within a polytunnel, (h) showing all tags without polytunnel obstruction, and (i) indicating how many tags in (h) were surveyed.}
\label{fig:groundtruth}
\end{figure}

In the strawberry polytunnels, a total of 960 tags were permanently mounted on structural poles in a diamond formation across eight complete tunnels (Fig. \ref{fig:strawberry_tags}). One tunnel features dense tag coverage at intervals of 2 to 3m to support high-precision ground truth validation, while the remaining seven tunnels employ sparse coverage at section boundaries for typical navigation tasks. In contrast, raspberry polytunnels required a more flexible installation strategy (Fig. \ref{fig:raspberry_tags}), tags were printed in pairs on cardboard and clipped to carefully selected locations to accommodate unpredictable plant growth that could otherwise occlude fixed mounting points.

\subsection{Total Station Tag Surveying}
We employed a Trimble S5 Total Station to establish millimetre-accurate ground truth positions for AprilTag landmarks. The total station measures coordinates of reflective targets affixed to tag-mounting locations with sub-centimetre precision. Due to visibility limitations within the polytunnel structure, we established a control network using prism points, enabling the total station to be repositioned across multiple known stations whilst maintaining coordinate system consistency. Detailed total station setup and tag placement procedures are illustrated in Fig.~\ref{fig:groundtruth}.

\subsection{Ground Truth Generation}
Poly-TagSLAM algorithm, which fuses LiDAR-inertial odometry with AprilTag detections, is adopted to generate the ground truth trajectory. Surveyed tag positions are incorporated as tight prior factors during an initial batch mapping run, after which ground truth trajectories for each sequence are obtained via incremental localisation with fixed tag positions. See~\cite{xu2025poly} for full algorithmic details.

\begin{figure}[h]
  \centering
  \includegraphics[width=0.9\linewidth]{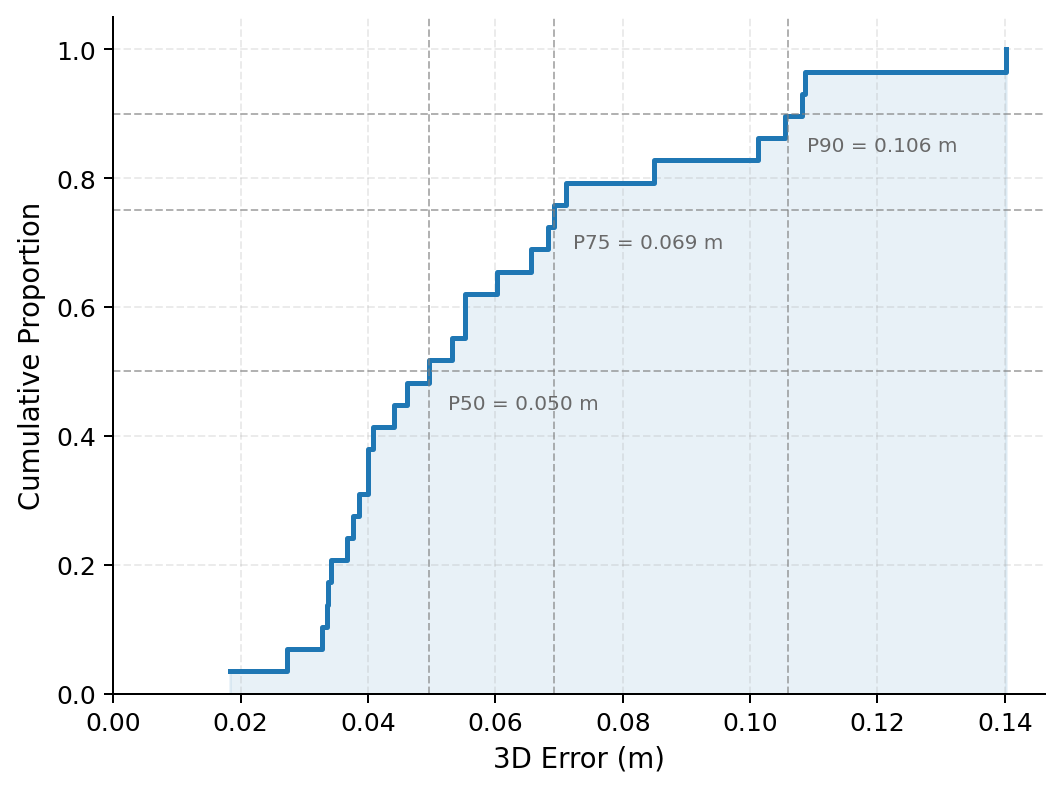}
  \caption{Cumulative distribution of localisation errors for held-out landmarks under LOO cross-validation.}
  \label{fig:loo}
\end{figure}

The accuracy of Poly-TagSLAM is validated through leave-one-out (LOO) cross-validation on 35 total-station-surveyed landmarks. For each surveyed landmark $L_i$, we optimise the factor graph using the remaining $N-1$ landmarks as tight priors ($\sigma = 1$cm). We then record and compare the 3D Euclidean error between the $L_i$ estimate and the surveyed position. Without survey constraints, raw Poly-TagSLAM estimates exhibit an RMSE of 14.1~cm, attributable to accumulated odometry drift over the 880~m trajectory. Incorporating all surveyed landmarks as tight priors ($\sigma = 1$~cm) reduces this to 7.3~cm RMSE. The LOO experiments demonstrate that this improvement is not an effect of over-fitting to the surveyed positions: when each landmark $L_i$ is estimated solely from odometry and neighbouring tag observations, held-out positions are recovered with a mean error of 5.9~cm and an RMSE of 6.6~cm, as summarised in Table~\ref{tab:landmark_accuracy}. Figure~\ref{fig:loo} shows the cumulative distribution of 3D errors across all held-out landmarks, with a median error of 5.0~cm. This compact distribution confirms that no landmark is systematically under-constrained. These results validate the suitability of Poly-TagSLAM as a ground truth source for SLAM evaluation in GPS-unreliable polytunnel environments.

\begin{table*}[htp]
\centering
\caption{Ground truth statistics for place recognition. Configuration used for all sequences: warm-up frames (skip first N) = 100, lower-bound frame gap (ignore immediate past) = 50, distance threshold = 10\,m, and top-1 closest neighbour.}

{\renewcommand{\arraystretch}{1.2} 
    \begin{tabular}{l
                    C{0.06\textwidth}
                    C{0.06\textwidth}
                    C{0.06\textwidth}
                    C{0.06\textwidth}
                    C{0.09\textwidth}
                    C{0.06\textwidth}
                    C{0.07\textwidth}
                    C{0.09\textwidth}
                    C{0.07\textwidth}
                    C{0.06\textwidth}}
    \toprule
    Sequence & Frames & Row Count & Query Pos. & Valid Loops & Loop Ratio (\%) & Min Dist (m) & Mean Dist (m) & Median Dist (m) & Max Dist (m) & Std Dist (m) \\
    \midrule
    Nov2025/Strawberry\_1 & 932 & 4 & 437 & 437 & 47 & 0.008 & 0.220 & 0.214 & 0.687 & 0.110 \\
    Jun2025/Strawberry\_1 & 1953 & 7 & 723 & 723 & 37 & 0.007 & 0.936 & 0.244 & 9.850 & 2.096 \\
    Jun2025/Raspberry\_1 & 340 & 4 & 181 & 181 & 53 & 0.017 & 0.225 & 0.168 & 1.929 & 0.267 \\
    Aug2025/Raspberry\_1 & 697 & 8 & 394 & 394 & 57 & 0.013 & 1.074 & 0.349 & 7.898 & 1.827 \\
    \bottomrule
    \end{tabular}
    }
    \label{tab:dataset:stats}
\end{table*}

\begin{table}[H]
\centering
\caption{Landmark Mapping Accuracy}
\label{tab:landmark_accuracy}
\begin{tabular}{lccc}
\toprule
\textbf{Configuration} & \textbf{RMSE (m)} & \textbf{ME (m)} & \textbf{Max error (m)} \\
\midrule
Poly-TagSLAM        & 0.1414 & 0.1269 & 0.2937 \\
Poly-TagSLAM + Total Station       & 0.0726 & 0.0351 & 0.2633 \\
LOO validation (held-out)  & 0.0655 & 0.0586 & 0.1403 \\
\bottomrule
\end{tabular}
\end{table}

The error observed under LOO validation is consistent with findings reported in similar works employing control-points for ground truth generation \cite{li2023whu, krishnan2025benchmarking}, and is largely attributable to the measurement error introduced by the human operator. For datasets where permanent tag infrastructure is not viable --- such as raspberry polytunnel sequences --- Poly-TagSLAM without survey constraints already achieves sufficient accuracy to serve as ground truth for the majority of SLAM algorithms, particularly visual SLAM systems. This configuration is therefore adopted as the ground truth source for all sequences in the dataset. 

For polytunnel sequences where permanent AprilTag installations are in place, we release the full tag maps estimated under survey-constrained optimisation, incorporating all 35 total-station-surveyed landmarks as tight priors. 

\subsection{Ground Truth For Place Recognition}
For multi-tunnel sequences, where the traversal spans multiple rows and revisits previously observed corridors, ground truth trajectory generation via Total Station or AprilTag-based localisation alone becomes increasingly impractical due to the cumulative complexity of tag placement and the extended spatial coverage required. In these sequences, ground truth is instead derived from \textbf{LIO-SAM}, whose single-tunnel accuracy is validated against our AprilTag-based reference system (see Table~\ref{tab:SLAM_sota_comparison}), providing confidence in its reliability as a reference for longer traversals. The resulting trajectories serve as ground truth specifically for place recognition evaluation in multi-tunnel sequences, where the primary challenge is the correct association of perceptually similar rows across revisits rather than per-frame localisation accuracy. Table~\ref{tab:dataset:stats} summarises the ground truth statistics for place recognition derived from LIO-SAM trajectories across the exampled four multi-tunnel sequences. These sequences are well-suited for evaluating place recognition, as their repeated row traversals guarantee frequent and geometrically verifiable loop closure opportunities.

\section{Dataset Use Example}
To demonstrate the utility of our dataset, we present two complementary comparative studies representative of the core challenges facing autonomous navigation in agricultural polytunnel environments. The first evaluates localisation performance through a systematic comparison of SOTA algorithms. The second examines place recognition, evaluating the ability to retrieve previously visited locations under the severe perceptual aliasing. These two studies are closely motivated by a key finding from our evaluation: whilst LiDAR and LiDAR-inertial methods are capable of achieving reliable localisation within a single tunnel traversal, performance degradations at larger scale operations are inevitable. Loop closure detection is the most viable mechanism for correcting such drift, yet it fundamentally depends on robust place recognition. Together, these studies illustrate how the dataset exposes a critical gap in the full localisation pipeline.

\begin{figure*}[!tp]
    \centering
    \begin{subfigure}[b]{0.42\textwidth}
    \centering
    \includegraphics[width=0.95\textwidth]{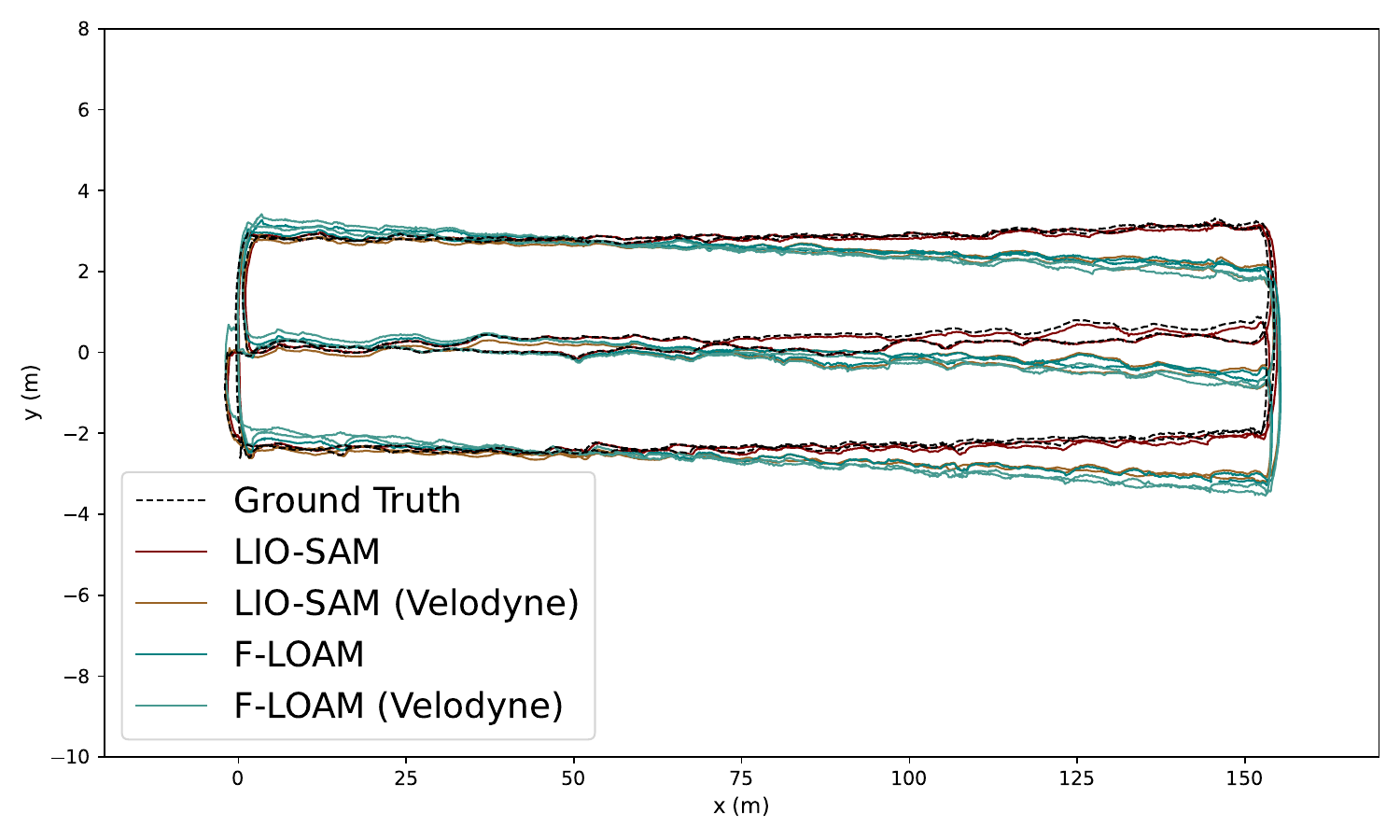}
    \caption{}
    \end{subfigure}
    \begin{subfigure}[b]{0.42\textwidth}
    \centering
    \includegraphics[width=0.95\textwidth]{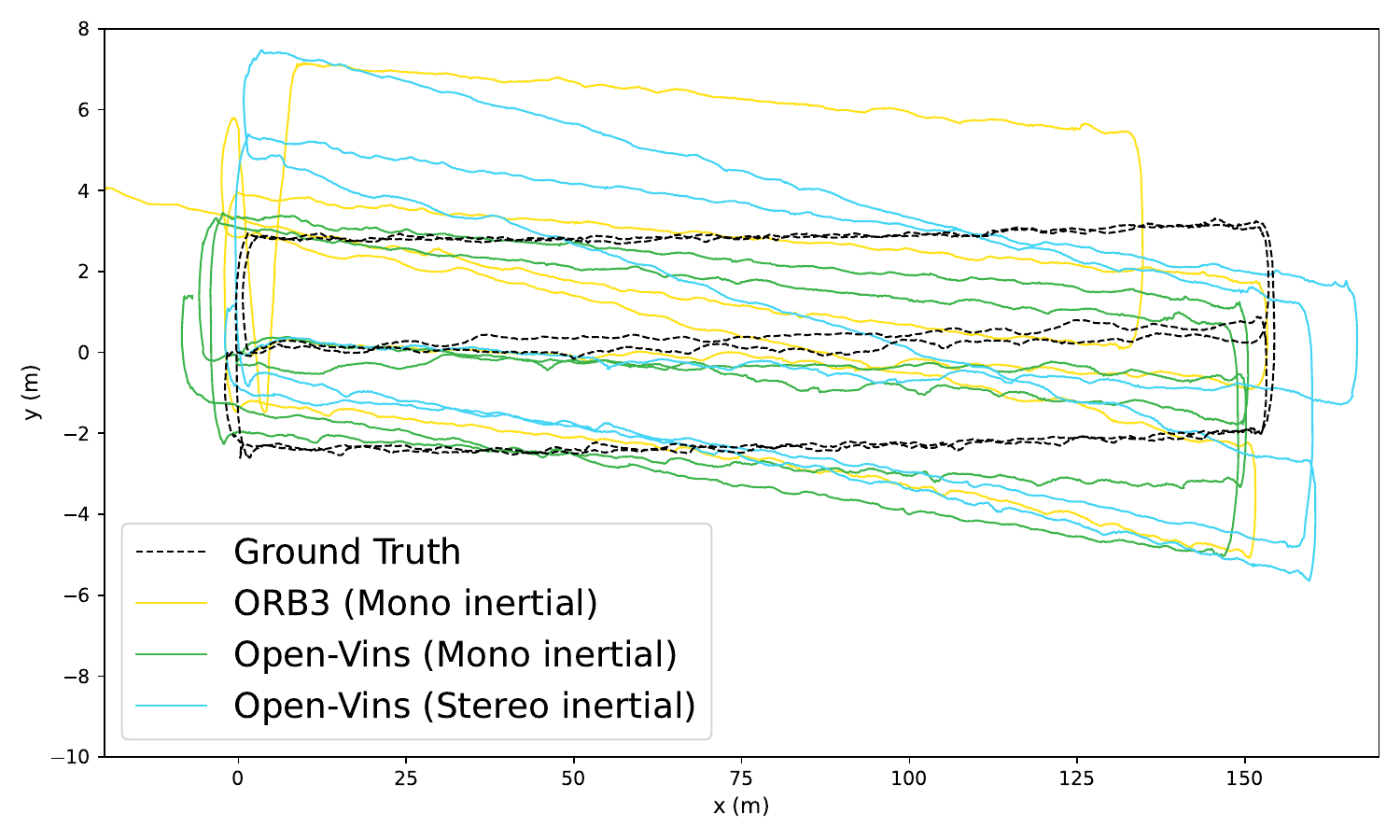}
    \caption{}
    \end{subfigure}
    \begin{subfigure}[b]{0.42\textwidth}
    \centering
    \includegraphics[width=0.95\textwidth]{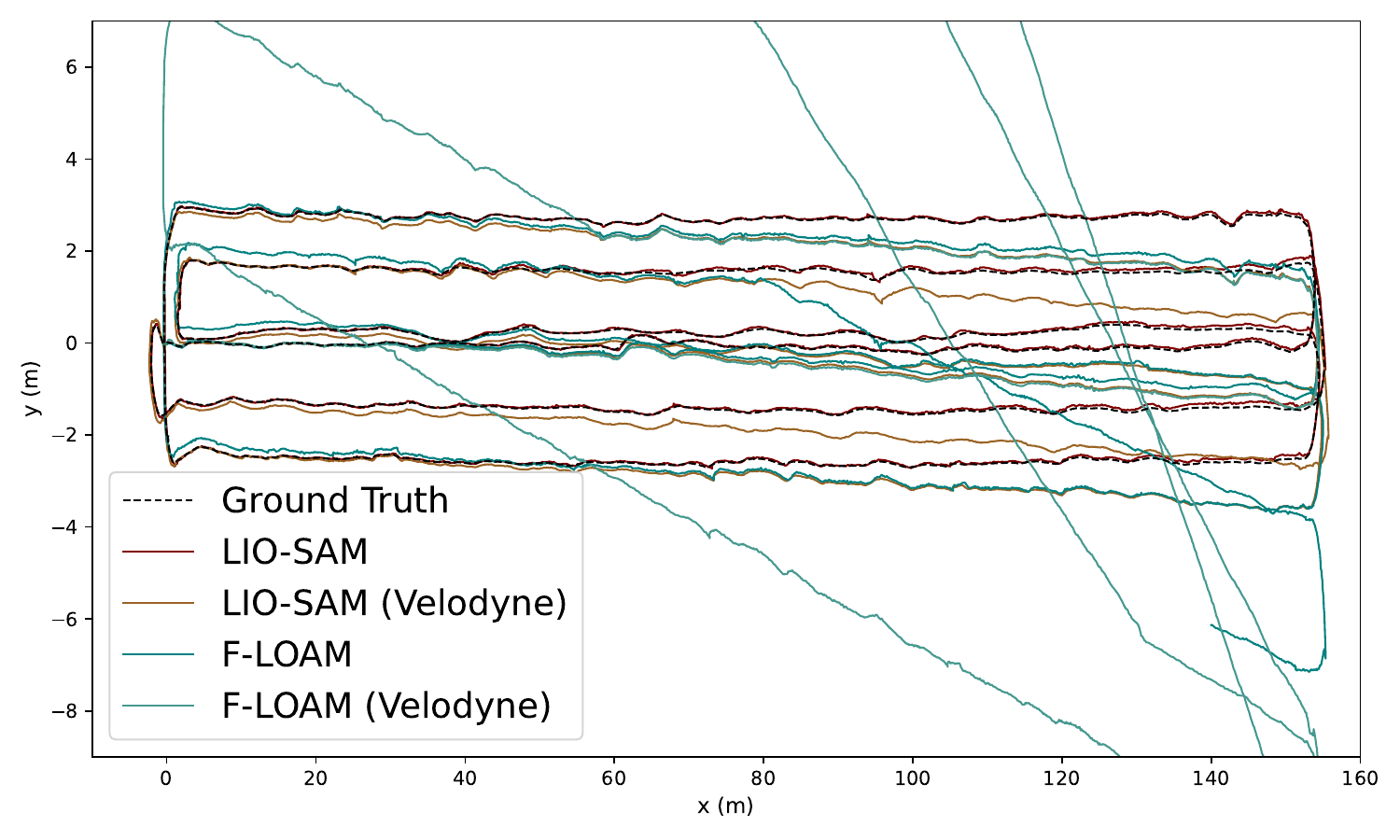}
    \caption{}
    \end{subfigure}
    \begin{subfigure}[b]{0.42\textwidth}
    \centering
    \includegraphics[width=0.95\textwidth]{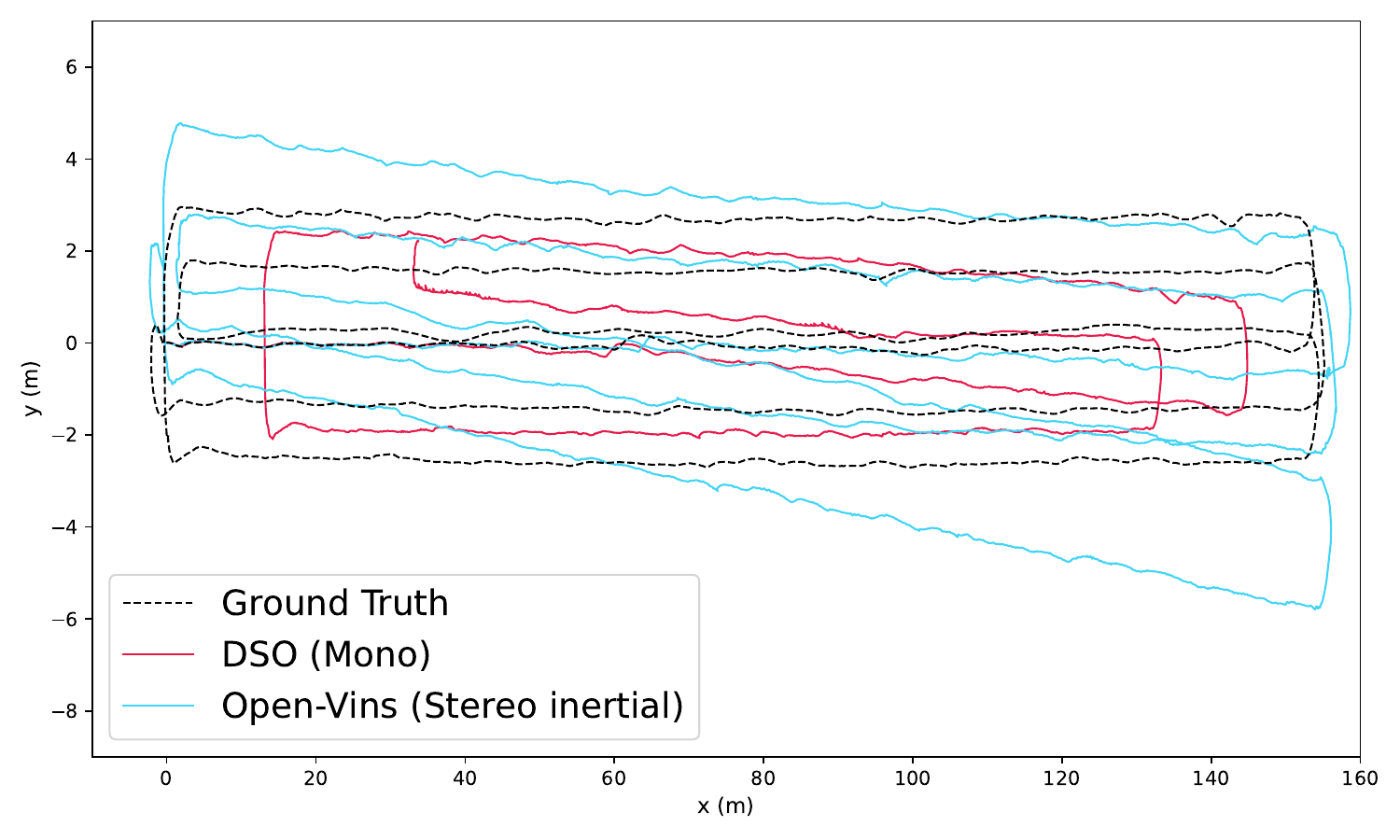}
    \caption{}
    \end{subfigure}
    \begin{subfigure}[b]{0.42\textwidth}
    \centering
    \includegraphics[width=0.95\textwidth]{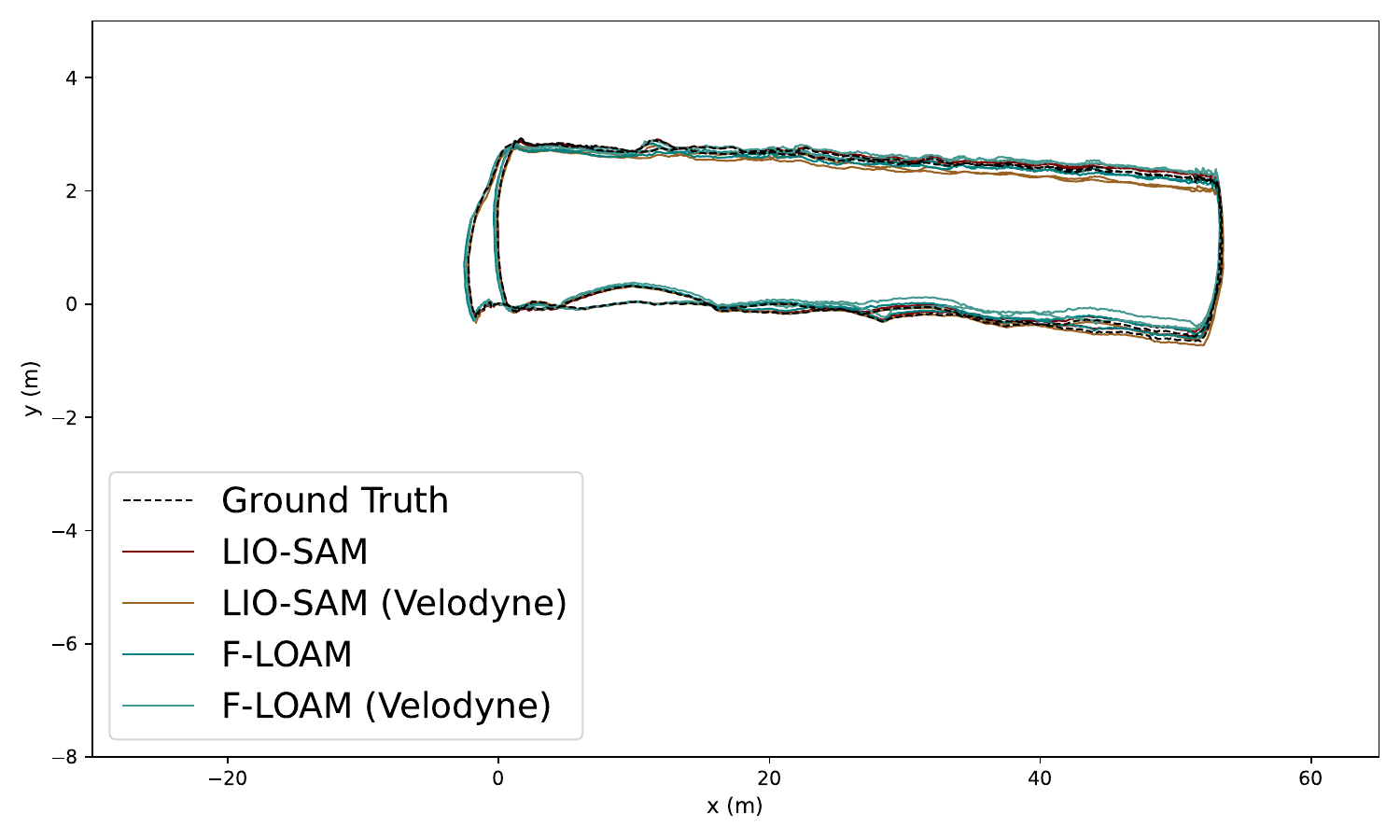}
    \caption{}
    \end{subfigure}
    \begin{subfigure}[b]{0.42\textwidth}
    \centering
    \includegraphics[width=0.95\textwidth]{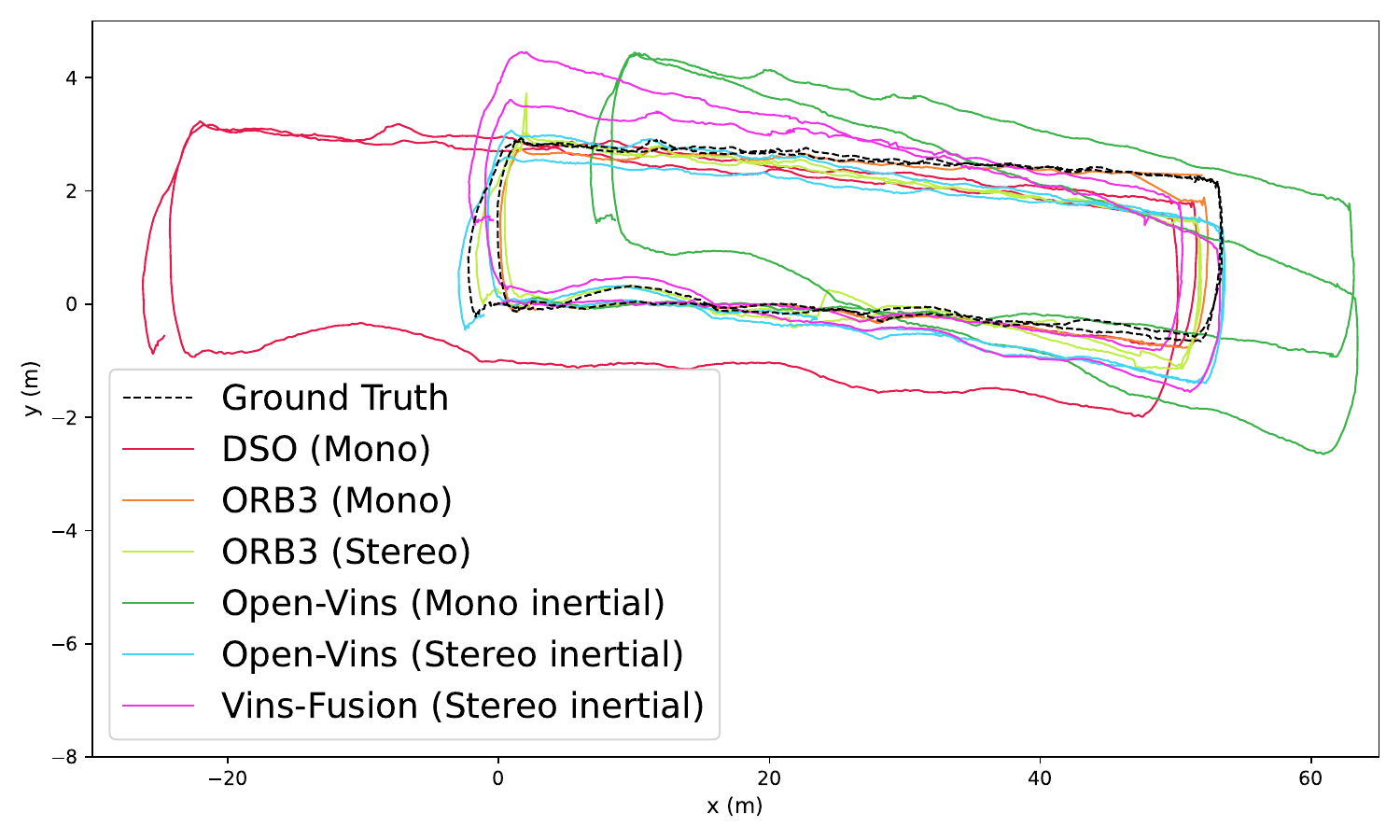}
    \caption{}
    \end{subfigure}
    \begin{subfigure}[b]{0.42\textwidth}
    \centering
    \includegraphics[width=0.95\textwidth]{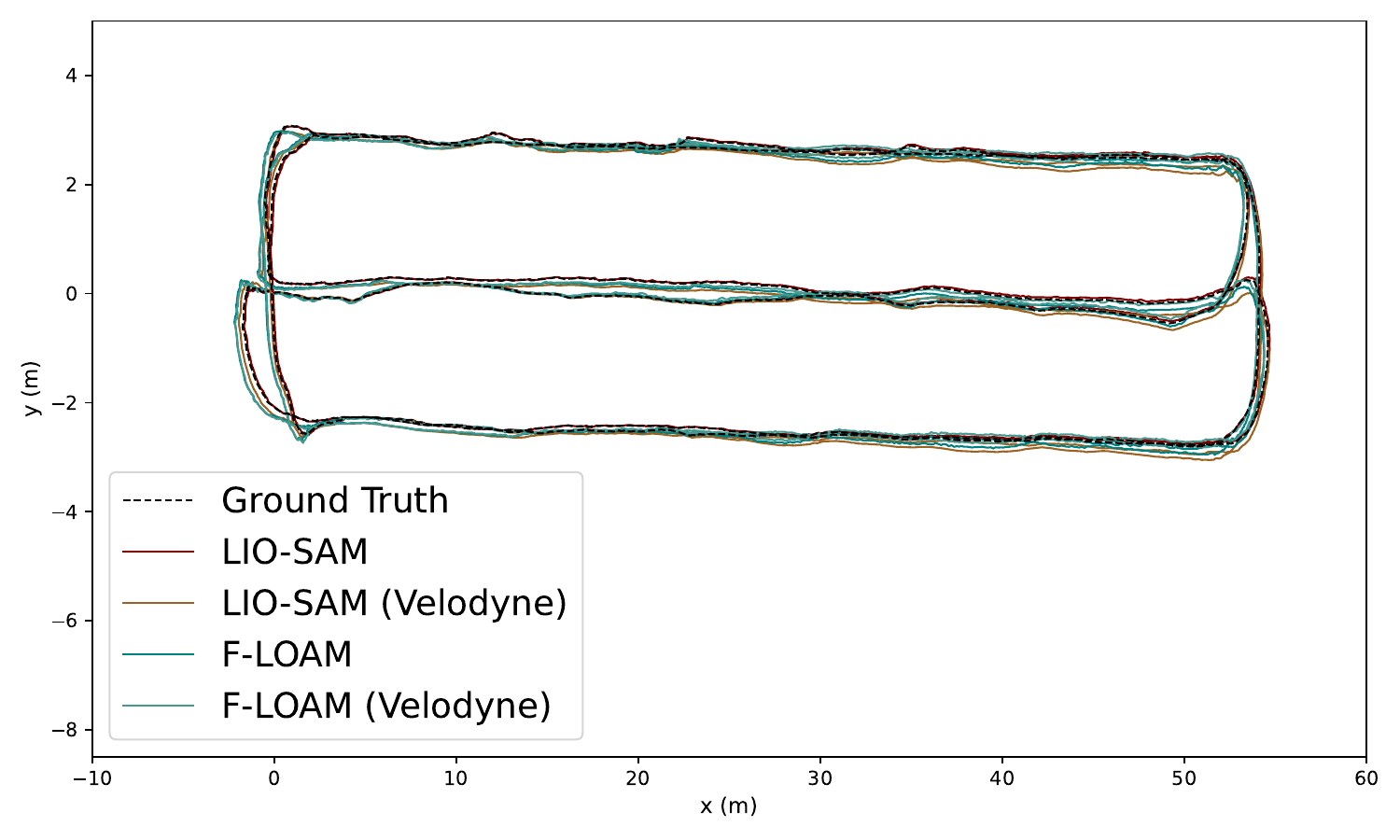}
    \caption{}
    \end{subfigure}
    \begin{subfigure}[b]{0.42\textwidth}
    \centering
    \includegraphics[width=0.95\textwidth]{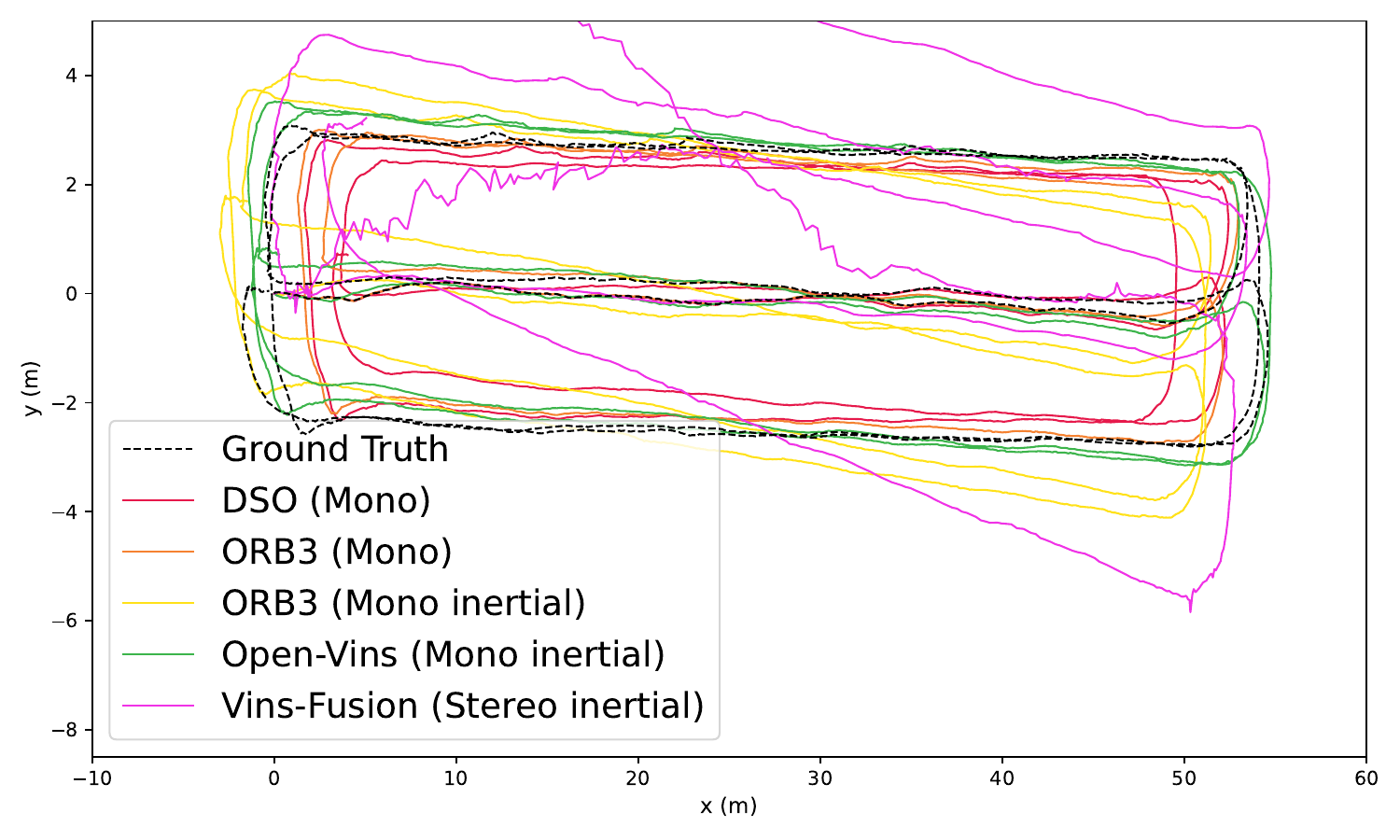}
    \caption{}
    \end{subfigure}
    \caption{Qualitative trajectory comparisons within the polytunnel. (a, b) sequence Feb2026$/$Strawberry$\_1$, (c, d) sequence Feb2026$/$Strawberry$\_2$ (e, f) sequence Feb2026$/$Strawberry$\_3$, and (g, h) sequence Feb2026$/$Strawberry$\_4$. These figures illustrate the performance and stability of SLAM algorithms with different designed robot trajectories, showing variations in drift, trajectory consistency, and overall accuracy}
    \label{fig:slamtrajectory}
\end{figure*}

\begin{table*}[tp]
\centering
\caption{Sequences specifications and SOTA SLAM performance comparison}
\label{tab:SLAM_sota_comparison}
\setlength{\tabcolsep}{4pt}

\begin{threeparttable}
{\renewcommand{\arraystretch}{1.3}  
\begin{tabular}{lcccccccc}
\toprule
\textbf{Stat/Sequence}(Feb2026/Strawberry\_) 
& \multicolumn{2}{c}{\textbf{1}} 
& \multicolumn{2}{c}{\textbf{2}} 
& \multicolumn{2}{c}{\textbf{3}} 
& \multicolumn{2}{c}{\textbf{4}} \\
\cmidrule(lr){2-3} \cmidrule(lr){4-5}
\cmidrule(lr){6-7} \cmidrule(lr){8-9}
& RPE (\%) & ATE (m)
& RPE (\%) & ATE (m)
& RPE (\%) & ATE (m)
& RPE (\%) & ATE (m) \\
\midrule
Duration (s)  & \multicolumn{2}{c}{876} 
              & \multicolumn{2}{c}{952} 
              & \multicolumn{2}{c}{242} 
              & \multicolumn{2}{c}{348} \\
Distance (m)  & \multicolumn{2}{c}{951.60} 
              & \multicolumn{2}{c}{940.90} 
              & \multicolumn{2}{c}{231.02} 
              & \multicolumn{2}{c}{350.37} \\
Size (GB)     & \multicolumn{2}{c}{37.1} 
              & \multicolumn{2}{c}{40.3} 
              & \multicolumn{2}{c}{10.2} 
              & \multicolumn{2}{c}{16.9} \\
\midrule
DSO (Mono)
& $\times$ & $\times$
& 32.07 & 10.2568
& 43.53  & 14.1276
& 22.84 & 2.2490 \\
ORB3 (Mono)
& $\times$ & $\times$
& $\times$ & $\times$
& 19.42 & 0.4920
& 11.16 & 1.5589 \\
ORB3 (Mono Inertial)
& 15.05 & 10.7027
& 18.27 & 10.7716
& 3.37 & 22.7727
& 5.63 & 2.1896 \\
ORB3 (Stereo)
& $\times$ & $\times$
& $\times$ & $\times$
& 8.00 & 0.9066
& 184.37 & 23.9799 \\
Open-Vins (Mono Inertial)
& 5.74 & 4.5396
& $\times$ & $\times$
& 15.60 & 9.0245
& 4.65 & 0.6653 \\
Open-Vins (Stereo Inertial)
& 9.50 & 5.9612
& 10.67 & 2.7691
& 5.60 & 0.6526
& $\times$ & $\times$  \\
VINS-Fusion (Mono Inertial)
& $\times$ & $\times$
& $\times$ & $\times$
& 41.29 & 17.5131
& $\times$ & $\times$ \\
VINS-Fusion (Stereo Inertial)
& $\times$ & $\times$
& $\times$ & $\times$
& 9.32 & 1.5009
& 20.99 & 3.9202 \\
\midrule
FLOAM
& 3.79 & 0.8587
& 10.64 & 1.6054
& 3.58 & 0.2879
& 4.10 & 0.2045 \\
FLOAM (Velodyne)
& 4.04 & 0.8763
& 25.80 & 8.9257
& 3.77 & 0.2460
& 4.22 & 0.2075 \\
LIO-SAM
& 1.64 & 0.4308
& 9.24 & 0.7189
& 0.64 & 0.0440
& 1.66 & 0.3850 \\
LIO-SAM (Velodyne)
& 3.01 & 0.8048
& 9.81 & 1.0408
& 2.71 & 0.1856
& 3.22 & 0.3442 \\
\bottomrule
\end{tabular}
}
\begin{tablenotes}
  \footnotesize
  \item[$\times$] Algorithm that did not produce a valid output trajectory
\end{tablenotes}

\end{threeparttable}
\end{table*}

\subsection{Test against SLAM Algorithms}
To demonstrate the utility of our dataset for robotic navigation research, we evaluate a representative set of SOTA SLAM approaches against our ground truth trajectories. Each method is assessed using Absolute Trajectory Error (ATE) and Relative Pose Error (RPE) \cite{sturm2012benchmark}, providing complementary measures of global consistency and local drift accumulation respectively. The selected sequences span a range of traversal lengths (231--952\,m) and trajectory complexities, from straightforward single-pass corridors to multi-row routes with frequent re-visits, enabling a thorough algorithm behaviour comparison under the unique conditions of agricultural polytunnel environments.

We evaluate methods across five modality categories, covering a broad spectrum of hardware configurations, to enable practitioners to make informed trade-offs between sensor cost, system complexity, and achievable navigation accuracy for their specific operational requirements when developing polytunnel localisation systems. Given the highly challenging nature of polytunnel environments for visual estimation, we deliberately include multiple candidate algorithms per modality to avoid drawing conclusions from a single method's failure, ensuring that observed limitations are attributable to the environment rather than to any individual implementation. For visual estimation, \textbf{DSO} \cite{engel2017direct} and \textbf{ORB-SLAM3} \cite{campos2021orb} cover direct and feature-based monocular paradigms, with ORB-SLAM3 also evaluated in monocular-inertial and stereo configurations. \textbf{OpenVINS} \cite{geneva2020openvins} and \textbf{VINS-Fusion} \cite{qin2018vins} are included for stereo-inertial estimation. For LiDAR-based methods, \textbf{FLOAM} \cite{wang2021} and \textbf{LIO-SAM} \cite{liosam2020shan} represent LiDAR-only and LiDAR-inertial configurations respectively, both evaluated with the Ouster OS0 and Velodyne VLP-16 to quantify the effect of point cloud density. Although tightly-coupled LiDAR-visual-inertial methods such as \textbf{LVI-SAM} \cite{lvisam2021shan} were also tested --- and our calibration and synchronisation quality was sufficient to run them --- none were able to complete any trajectory, so their results are not shown.

The results, detailed in Table~\ref{tab:SLAM_sota_comparison}, highlight the fundamental challenges polytunnel environments pose to existing localisation methods. Visual and visual-inertial approaches struggle severely: the repetitive rows, limited lighting conditions, and low-texture tunnel structure provide insufficient features for reliable estimation, with the majority of monocular methods failing and even tightly-coupled stereo-inertial configurations degrading on several sequences. LiDAR-based methods are more robust, though FLOAM's drift accumulation on longer traversals, illustrates that LiDAR-only odometry without inertial support remains vulnerable to the long, geometrically degenerate corridors characteristic of polytunnels. Across all LiDAR methods, the Ouster OS0 consistently outperforms the VLP-16, with the gap widening on harder sequences, underscoring the sensitivity of scan-matching to point cloud density in this environment.

\begin{figure*}[thb]
 \centering
    \begin{subfigure}[b]{0.49\columnwidth}
    \centering
\includegraphics[width=\columnwidth, trim={0.1cm  1.0cm 2.5cm  3.0cm},clip]{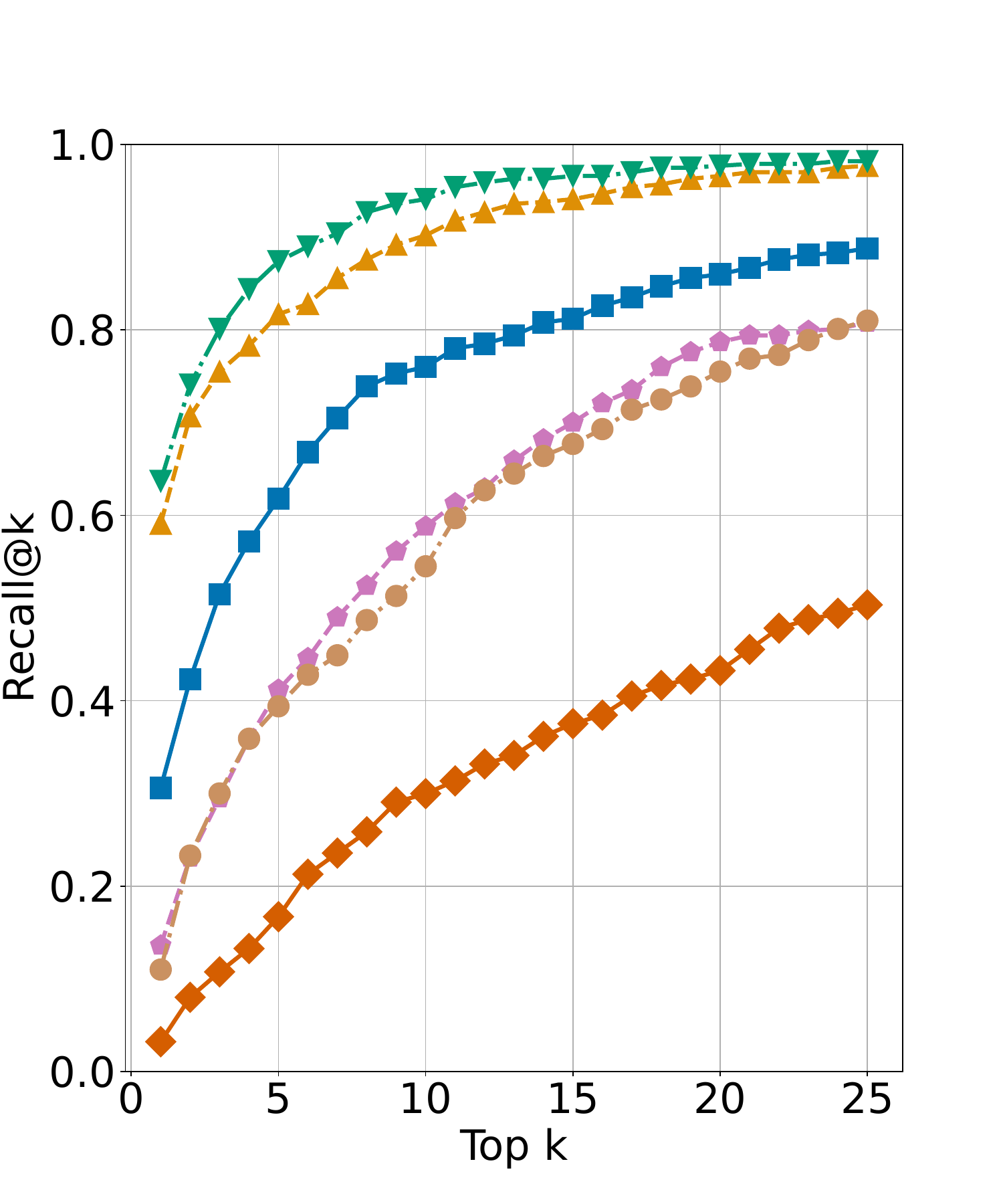}
    \caption{}
  \end{subfigure}
  \hfill
  \begin{subfigure}[b]{0.49\columnwidth}
    \centering
    \includegraphics[width=\columnwidth, trim={0.1cm  1.0cm 2.5cm  3.0cm},clip]{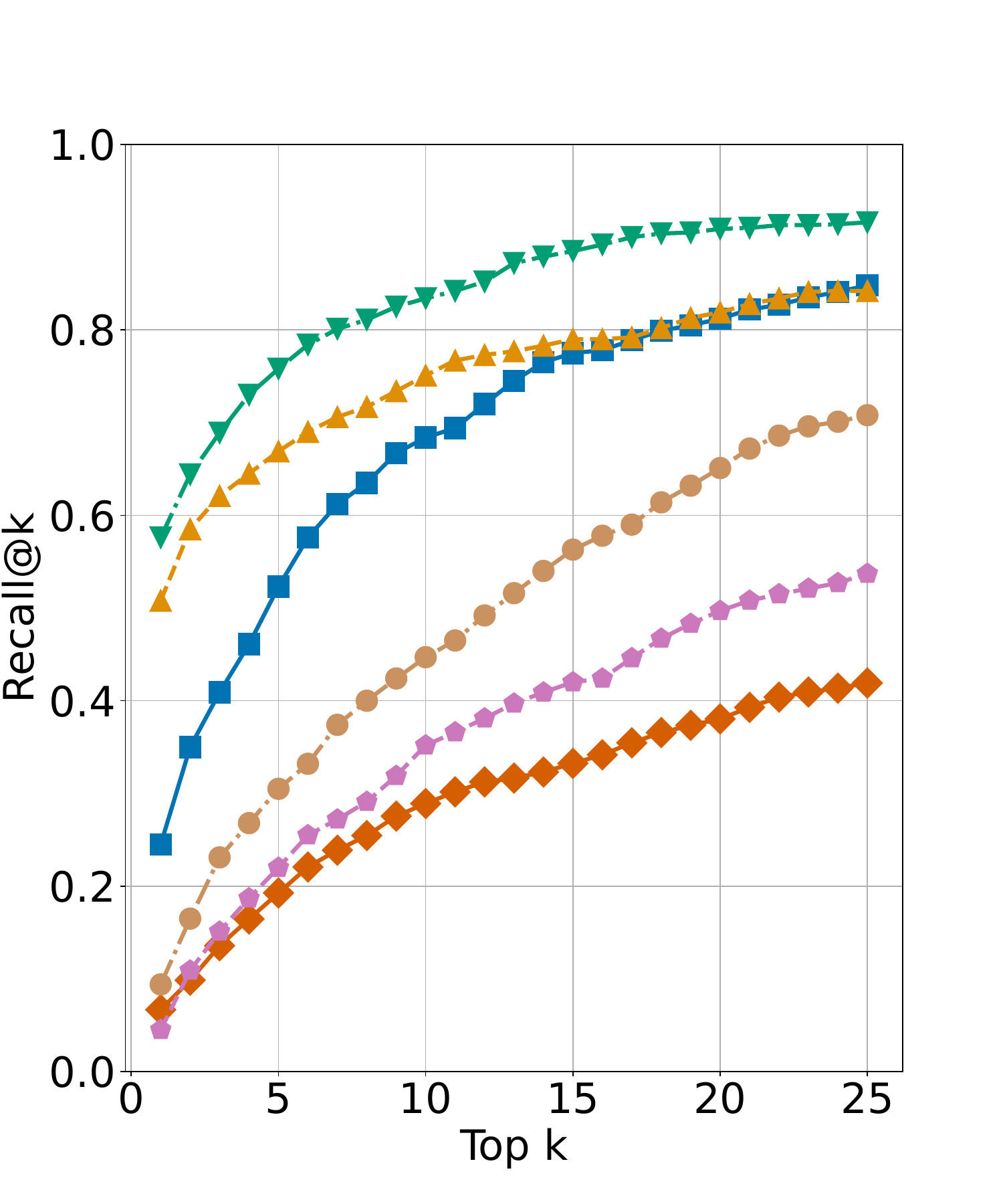}
    \caption{}
  \end{subfigure}
  \hfill
  \begin{subfigure}[b]{0.49\columnwidth}
    \centering
\includegraphics[width=\columnwidth, trim={0.1cm  1.0cm 2.5cm  3.0cm},clip]{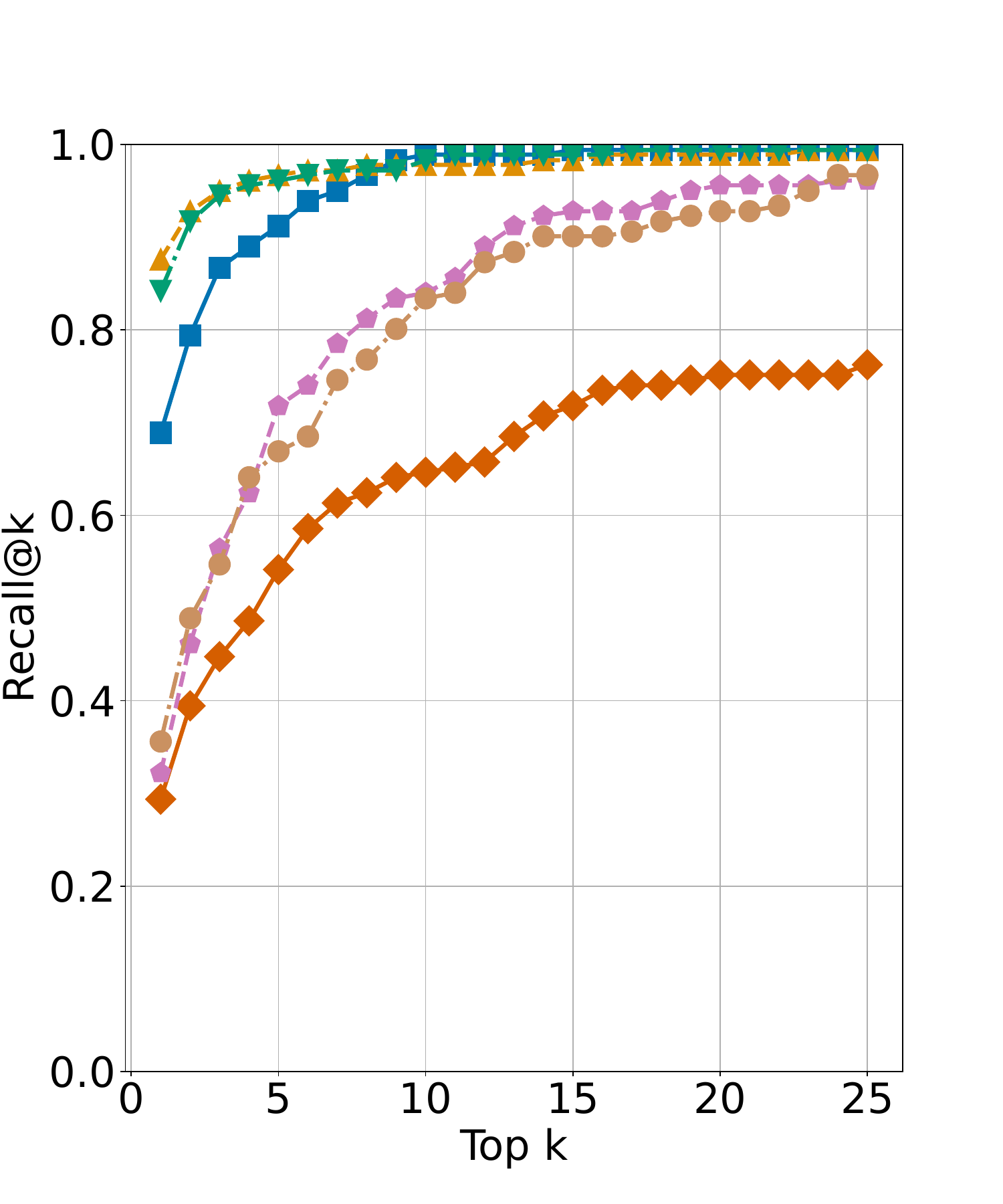}
    \caption{}
  \end{subfigure}
    \begin{subfigure}[b]{0.49\columnwidth}
    \centering
\includegraphics[width=\columnwidth, trim={0.1cm  1.0cm 2.5cm  3.0cm},clip]{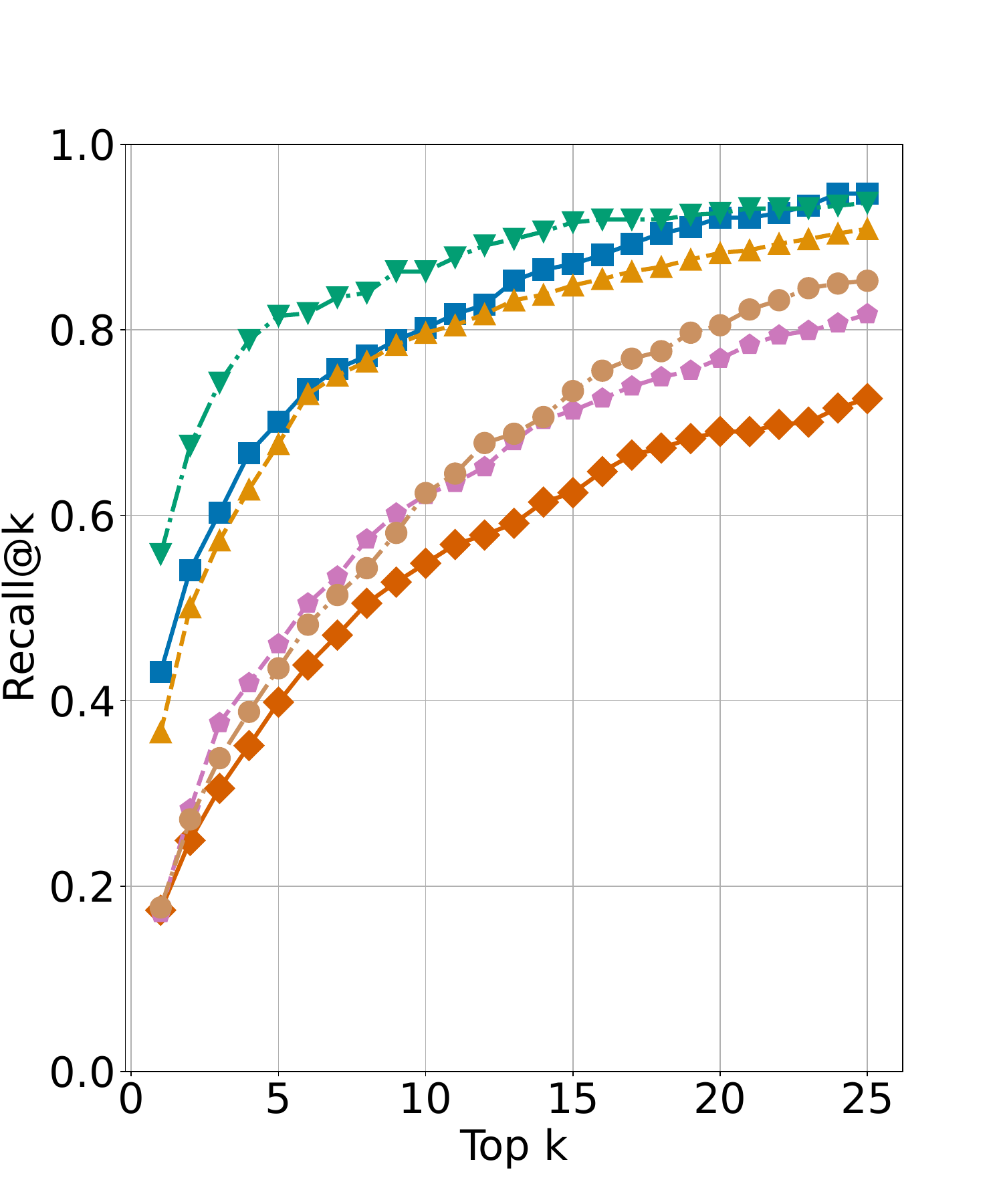}
    \caption{}
  \end{subfigure}
  \begin{subfigure}[b]{1\textwidth}
    \centering
\includegraphics[width=\textwidth, trim={3.0cm  0.0cm 3.5cm  0.0cm},clip]{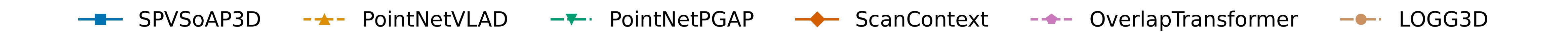}
  \end{subfigure}
  \caption{Retrieval performance for the top-25 candidates within a 10\,m range} 
  \label{fig:pr:top25}
\end{figure*}

\begin{table*}[thb]
\centering
\caption{Retrieval performance, using recall, of the top-1 retrieved candidate for various threshold range.}
{\renewcommand{\arraystretch}{1.3} 
\begin{tabular}{l|ccc|ccc|ccc|ccc}
\toprule
 & \multicolumn{3}{c}{Jun2025/Strawberry\_1} & \multicolumn{3}{c}{Nov2025/Strawberry\_1} & \multicolumn{3}{c}{Jun2025/Raspberry\_1} & \multicolumn{3}{c}{Aug2025/Raspberry\_1} \\
 & @5m & @10m & @25m & @5m & @10m & @25m & @5m & @10m & @25m & @5m & @10m & @25m \\
\midrule
SPVSoAP3D\cite{10802603}  & 0.201 & 0.245 & 0.356 & 0.251 & 0.306 & 0.384 & 0.633 & 0.689 & 0.768 & 0.309 & 0.431 & 0.519 \\
PointNetVLAD\cite{angelina2018pointnetvlad} & 0.452 & 0.508 & 0.598 & 0.531 & 0.591 & 0.660 & 0.836 & 0.876 & 0.944 & 0.212 & 0.366 & 0.428 \\
PointNetPGAP\cite{10706020} & 0.504 & 0.576 & 0.668 & 0.508 & 0.637 & 0.692 & 0.797 & 0.842 & 0.915 & 0.430 & 0.558 & 0.596 \\
LoGG3D-Net\cite{9811753} & 0.068 & 0.094 & 0.148 & 0.083 & 0.110 & 0.182 & 0.282 & 0.356 & 0.554 & 0.091 & 0.177 & 0.265 \\
OverlapTransformer\cite{9785497} & 0.028 & 0.045 & 0.076 & 0.094 & 0.136 & 0.209 & 0.254 & 0.322 & 0.395 & 0.064 & 0.171 & 0.227 \\
ScanContext\cite{gkim-2018-iros} & 0.038 & 0.067 & 0.123 & 0.016 & 0.032 & 0.129 & 0.062 & 0.294 & 0.401 & 0.118 & 0.174 & 0.242 \\
\bottomrule
\end{tabular}
}
\label{tab:pr:top1}
\end{table*}

\begin{figure*}[htb]
 \centering
    \centering
    \includegraphics[width=1\textwidth, trim={0.0cm  0.0cm 0.0cm  0.0cm},clip]{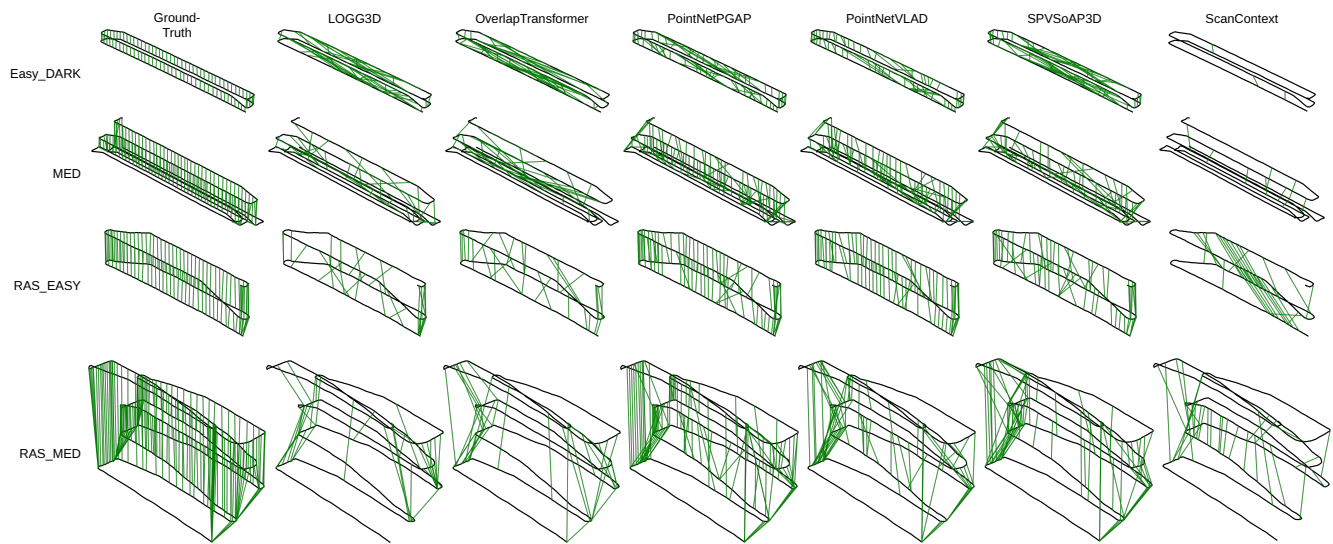}
  \caption{Illustration of the top-1 true positive predictions within a 10\,m range. For clear visualization only a fraction of the true positives are plotted.}
  \label{fig:pr:top1}
\end{figure*}

\subsection{Test against Place Recognition Algorithms}

To assess the difficulty of polytunnel environments for global re-localisation, we benchmark a set of SOTA place recognition methods on the proposed dataset. Place recognition (PR) addresses the problem of global re-localisation by identifying previously visited locations using perceptual information. 

The evaluation was conducted using a cross-validation protocol, where SOTA  learning-based LiDAR place recognition approaches such as  PointNetVLAD~\cite{angelina2018pointnetvlad}, SPVSoAP3D~\cite{10802603}, PointNetPGAP~\cite{10706020},  LoGG3D-Net~\cite{9811753}  and  OverlapTransformer~\cite{9785497}, were trained from scratch on the Horto3DLM dataset~\cite{10706020,10802603} and evaluated directly on four PolyTunnel sequences: $Nov2025/Strawberry\_1$, $Jun2025/Strawberry\_1$, $Jun2025/Raspberry\_1$, and $Aug2025/Raspberry\_1$.  Scan Context (SC)~\cite{gkim-2018-iros}, which is a handcrafted approach,  was evaluated directly on the four sequences. Table~\ref{tab:dataset:stats} presents the ground truth statistics of the selected sequences.  The performance is reported using Recall@K. A retrieval is considered correct if the Euclidean distance between the ground truth poses of the query and retrieved scan is below this threshold (in metres).

The quantitative retrieval performance is reported in  Table~\ref{tab:pr:top1}, with Recall@K curves illustrated in Fig.~\ref{fig:pr:top25} and qualitative examples shown in Fig.~\ref{fig:pr:top1}. Across all sequences, the results confirm that commercial polytunnels constitute a highly challenging benchmark for LiDAR-based place recognition. Under the strict 10\,m localisation threshold and Top-1 retrieval setting, performance remains substantially lower than typically reported on urban driving or structured outdoor datasets~\cite{10706020}, highlighting the impact of severe structural repetition and perceptual aliasing.

Learning-based global descriptor methods consistently outperform handcrafted approaches across all difficulty levels. In particular, PointNetPGAP~\cite{10706020} and PointNetVLAD~\cite{angelina2018pointnetvlad} achieve the highest Top-1 recall at 10\,m and 25\,m thresholds in most sequences, indicating that learned feature aggregation improves discriminative capacity under repetitive row geometries.

In contrast, Scan Context~\cite{gkim-2018-iros} exhibits noticeable degradation. As Scan Context relies on a fixed polar projection, crop rows and repeated metallic supports produce highly similar radial patterns across spatially distinct locations. This leads to descriptor ambiguity and increased false positives, particularly between adjacent rows and symmetric tunnel segments.

Importantly, none of the evaluated methods achieves uniformly high recall under the strict Top-1, 10\,m threshold. This reflects the difficulty of polytunnel environments, where long, similar corridors constrain global distinctiveness. While learning-based aggregation partially mitigates descriptor collapse, geometry-only representations remain vulnerable to aliasing in highly structured cultivation systems.

\section{Conclusion}
We present HortiMulti, the first comprehensive multi-sensor, multi-season localisation and mapping dataset specifically targeting commercial horticultural polytunnels, addressing a critical gap in the agricultural robotics community. The dataset spans strawberry and raspberry polytunnel environments across a full growing season, capturing the seasonal appearance changes, perceptual aliasing, dynamic foliage, and GNSS-unreliable conditions that define this application domain. A rigorous survey-grade ground truth pipeline combining Total Station surveying with Poly-TagSLAM achieves sub 7cm landmark accuracy, validated through leave-one-out cross-validation. Benchmarking results across a representative selection of LiDAR, visual, and multimodal SLAM algorithms reveal that current methods — even SOTA LiDAR-inertial systems — remain vulnerable to geometric degeneracy and drift accumulation in the repetitive row-corridor geometry of polytunnels, while all evaluated visual methods struggle significantly. Place recognition results further confirm that polytunnels expose fundamental limitations in both handcrafted and learning-based retrieval methods, with no evaluated approach achieving reliable recall under strict localisation thresholds. Together, these findings demonstrate that HortiMulti is a challenging and practically motivated benchmark that we hope will catalyse the development of localisation, mapping, and perception algorithms capable of sustaining robust operation in the demanding conditions of commercial horticulture.

Looking ahead, HortiMulti is intended as a continuously expanding resource. We plan to release additional sequences spanning further seasonal cycles. We also welcome community contributions of benchmarking, annotations, and other implementations, with the aim of establishing HortiMulti as a shared reference tool for the horticultural robotics community.


\bibliographystyle{IEEEtran}
\bibliography{hortislam_refs}

\vfill\pagebreak

\end{document}